\pdfoutput=1
\documentclass[11pt]{article}

\usepackage[final]{acl}

\usepackage{times}
\usepackage{latexsym}
\usepackage{amsmath}
\usepackage[T1]{fontenc}
\usepackage[utf8]{inputenc}

\usepackage{microtype}

\usepackage{inconsolata}

\usepackage{graphicx}

\usepackage{xcolor}
\usepackage{svg}
\usepackage{tikz}
\usepackage{verbatim}
\usepackage{booktabs}
\usepackage{multirow}
\usepackage{float}
\usepackage{algorithm}
\usepackage{algpseudocode}
\usepackage{caption}
\usepackage{makecell}
\usepackage{colortbl}
\usepackage{enumitem}

\title{LLM as Effective Streaming Processor: Bridging Streaming-Batch Mismatches with Group Position Encoding}

\author{
 \textbf{Junlong Tong\textsuperscript{1,2,3}},
 \textbf{Jinlan Fu\textsuperscript{4}},
 \textbf{Zixuan Lin\textsuperscript{5}},
 \textbf{Yingqi Fan\textsuperscript{3}},
 \\
\textbf{Anhao Zhao\textsuperscript{3}},
 \textbf{Hui Su\textsuperscript{6}},
 \textbf{Xiaoyu Shen\textsuperscript{2,3}\thanks{Corresponding author}}
\\
 \textsuperscript{1}Shanghai Jiao Tong University\\
 \textsuperscript{2}Ningbo Key Laboratory of Spatial Intelligence and Digital Derivative\\
 \textsuperscript{3}Institute of Digital Twin, EIT \quad \textsuperscript{4}National University of Singapore \\
 \textsuperscript{5}University of Science and Technology of China\quad
 \textsuperscript{6}Meituan Inc.
\\
\texttt{jl-tong@sjtu.edu.cn}~~~~~\texttt{xyshen@eitech.edu.cn}
}

\begin{document}
\maketitle

\begin{abstract}
%% version-3
Large Language Models (LLMs) are primarily designed for batch processing. Existing methods for adapting LLMs to streaming rely either on expensive re-encoding or specialized architectures with limited scalability.
This work identifies three key mismatches in adapting batch-oriented LLMs to streaming: (1) input-attention, (2) output-attention, and (3) position-ID mismatches.
While it is commonly assumed that the latter two mismatches require frequent re-encoding, our analysis reveals that only the input-attention mismatch significantly impacts performance, indicating re-encoding outputs is largely unnecessary.
% To reconcile this observation with the common assumption,
To better understand this discrepancy with the common assumption,
we provide the first comprehensive analysis of the impact of position encoding on LLMs in streaming, showing that preserving relative positions within source and target contexts is more critical than maintaining absolute order.
Motivated by the above analysis, we introduce a group position encoding paradigm built on batch architectures to enhance consistency between streaming and batch modes. 
Extensive experiments on cross-lingual and cross-modal tasks demonstrate that our method outperforms existing approaches. 
Our method requires no architectural modifications, exhibits strong generalization in both streaming and batch modes.
% and effectively supports real-world streaming applications.
The code is available at repository \href{https://github.com/EIT-NLP/StreamingLLM}{\nolinkurl{https://github.com/EIT-NLP/StreamingLLM}}.

\end{abstract}
\section{Introduction}
\label{Introduction}

Large language models (LLMs) have revolutionized a multitude of tasks \cite{zhang2023video,liu2024visual,chu2023qwen,kojima2022large,kocmi2023large}. However, research on LLMs has largely focused on \textit{batch-processing}, where the entire input is processed at once \cite{zhao2023survey}. In contrast, human cognition operates incrementally, interpreting information as it arrives—a capability essential for real-time decision-making, interactive dialogue, and other latency-sensitive applications \cite{gonzalez2003instance,altmann2009incrementality}. \emph{Bridging this gap between batch-oriented LLMs and streaming-aware processing} is vital for unlocking their potential in dynamic, real-world scenarios.
% , such as simultaneous translation \cite{}, live speech transcription, and live streaming video 

\begin{figure}[t]
    \centering
    \includegraphics[width=\linewidth]{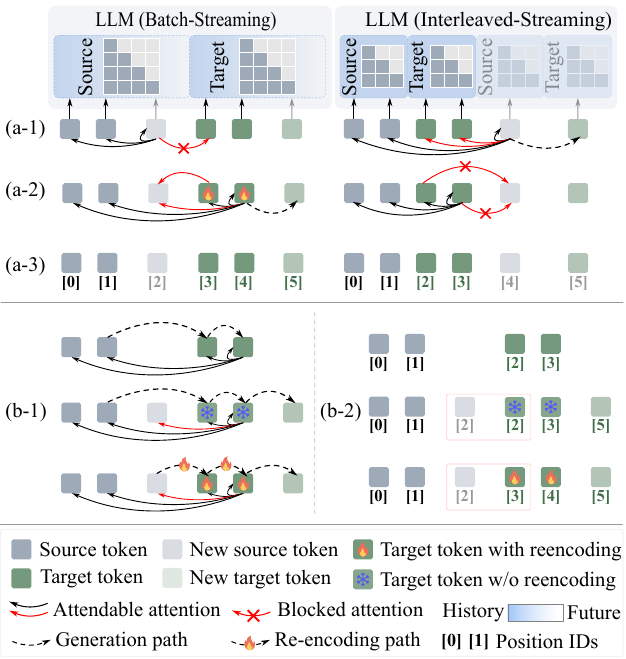}
    \caption{
    %verion-3
        Two streaming paradigms of LLMs: (a) Batch-streaming simulates batch-processing, while interleaved-streaming encodes streaming data in arrival order. (a-1) \textit{Input-Attention Mismatch}: Whether the source tokens can attend to the target tokens. (a-2) \textit{Output-Attention Mismatch}: Whether the target tokens can attend to the new source token. (a-3) \textit{Position-ID Mismatch}: Whether the position IDs reflect the actual token order. (b) Batch-streaming relies on (b-1) KV cache re-encoding and (b-2) position re-encoding to simulate batch-processing.
    }
    \label{Fig:intro}
\end{figure}

A naive strategy to adapt LLMs for streaming involves iteratively re-encoding both new inputs and prior outputs with each incoming data segment \cite{agostinelli-etal-2024-simul, wang-etal-2024-simultaneous,guo2024sillm,koshkin-etal-2024-transllama}, as illustrated in Figure~\ref{Fig:intro}(b). While this \textit{batch-streaming} paradigm preserves compatibility with batch-processing architectures, it introduces prohibitive computational costs.  Existing efforts to optimize LLMs for streaming data typically fall into two categories: (1) \emph{Directly encoding streaming data in arrival order} \cite{du2024cosyvoice,yang2024interleaved}, an \textit{interleaved-streaming} paradigm, which introduces structural mismatches with batch-processing setups used in pre-training and degrades performance;
(2) \emph{Designing entirely new architectures tailored to the streaming mode} \cite{guo-etal-2024-decoder,tsunoo2024decoder,chen2024streaming}, which is costly, lacks scalability, and fails to fully leverage pre-trained LLM capabilities.
Furthermore, existing methods lack rigorous analysis of the fundamental discrepancies between batch and streaming processing modes.

This work tackles these limitations by identifying three key mismatches in adapting batch-oriented LLMs to streaming, as shown in Figure~\ref{Fig:intro}:
\begin{itemize}[left=3pt]
    \item \textbf{Input-Attention Mismatch}:Batch-streaming confines input tokens to attending only prior inputs, whereas interleaved-streaming permits attention to previously decoded outputs.
    \item \textbf{Output-Attention Mismatch}:Batch-streaming allows decoded output tokens to attending to all received input tokens by\textit{ KV cache re-encoding}, while interleaved-streaming mode limits each output token's attention to the subset of inputs available at decoding time.
    \item \textbf{Position-ID Mismatch}: 
    Batch-streaming relies on \textit{position re-encoding}, assigning contiguous position IDs to inputs followed by outputs, whereas interleaved-streaming processes alternate between inputs and outputs incrementally, resulting in discontinuous positional ids that disrupt sequential coherence.
\end{itemize}

Building on the identification of these mismatches, we systematically studied their effects on LLM performance. Our analysis revealed that input-attention mismatch does affect streaming model performance.
In contrast, output-attention and position-ID mismatches have negligible effects.
A common assumption is that streaming models require re-encoding of previously generated content to mitigate \textit{token position inconsistencies} arising from the incremental nature of streaming setting \cite{raffel-etal-2024-simultaneous,guo-etal-2024-decoder,he2024let}, as shown in Figure~\ref{Fig:intro}(b).
However, our empirical findings do not support this hypothesis. Instead, we observe that re-encoding the output is not necessary \footnote{We clarify that re-encoding the target tokens is solely for refining the generation of the latest token without altering previously generated content.}.
This discrepancy with the common assumption raises a fundamental question: 
\textbf{How does position encoding impact LLMs in streaming scenarios? And how should we design appropriate position encoding for streaming LLMs?}

Existing research on positional encoding in LLMs has largely focused on static scenarios \cite{likhomanenko2021cape,haviv2022transformer,kazemnejad2024impact}, while its role in streaming scenarios remains underexplored.
We conducted a more in-depth analysis to further explore the impact of position encoding on streaming models. Experimental results reveal that the absolute positional order of tokens has a negligible effect on model performance in streaming tasks. However, maintaining the internal relative order within the source and target sequences is significantly more important. 
Based on the findings, we propose a grouped position encoding streaming paradigm built on batch architectures(\textit{group-streaming}), which groups input and output position ids to enable more consistent processing with the batch model. This strategy is not only computationally efficient but also generalizable across different tasks and model architectures. We validated its effectiveness on cross-lingual (machine translation) and cross-modal (automatic speech recognition) tasks, demonstrating that it significantly outperforms existing solutions, 
including LLMs with more complex streaming-optimized architectures.

The main contributions of this study can be summarized as: (1) We systematically analyze the mismatches between batch and streaming processing in LLMs, providing deep insights into key factors affecting their adaptation to streaming. Contrary to mainstream assumptions, our experiments reveal that position disorder is not the primary factor affecting LLM streaming performance. (2) We conduct the first comprehensive study on the impact of position encoding in streaming scenarios, demonstrating that absolute positional order is unnecessary, while maintaining relative order within source and target contexts is more critical. (3) We introduce a \textit{group streaming paradigm} for streaming LLMs. This method imposes no architectural constraints on batch-processing LLMs, allowing seamless application to any pre-trained LLM while ensuring high scalability and adaptability to various real-world streaming tasks.

\section{Streaming-Batch Mismatches}
\label{sec:mismatches}

LLMs are pre-trained in a batch-processing paradigm, where the entire input sequence $\mathbf{X} = [x_1, \dots, x_n]$ is processed simultaneously to generate the output sequence $\mathbf{Y} = [y_1, \dots, y_m]$. This paradigm assumes full input availability, allowing both self-attention and cross-attention mechanisms to operate over complete sequences. In contrast, streaming tasks require incremental processing, where inputs and outputs arrive and are processed in an interleaved manner over time. At any time step $t$, the model only has access to a partial input sequence $\mathbf{X}_t = [x_1, \dots, x_t]$ and generates a corresponding partial output sequence $\mathbf{Y}_{t'} = [y_1, \dots, y_{t'}]$. This shift from batch to streaming introduces three key mismatches:

% \subsection{Input-Attention Mismatch}
\paragraph{Input-Attention Mismatch}
% In batch-streaming mode, the self-attention mechanism enforces a strict ordering where each input token $x_i$ can only attend to prior inputs $\mathbf{X}_{<i}$. This is typically expressed as:
In batch-streaming mode, self-attention enforces a strict ordering, where each input token $x_i$ can only attend to prior inputs $\mathbf{X}_{<i}$. This is typically expressed as:
\begin{equation}
    h_i = \text{SelfAttention}(x_i, \mathbf{X}_{<i}),
\end{equation}
% where $h_i$ is the hidden representation of $x_i$. However, in interleaved-streaming mode, since outputs are generated incrementally, previously decoded outputs $\mathbf{Y}_{<t'}$ become available and are included in the attention context:
where $h_i$ is the hidden representation of $x_i$. However, in interleaved-streaming mode, as outputs are generated incrementally, previously decoded outputs $\mathbf{Y}_{<t'}$ become available and are included in the attention context:
\begin{equation}
    h_i^\text{interleaved} = \text{SelfAttention}(x_i, \mathbf{X}_{<i} \cup \mathbf{Y}_{<t'}).
\end{equation}
% This disrupts the model’s pre-trained assumptions, as input tokens in batch mode never attend to output tokens, potentially leading to degraded performance.
This disrupts the model’s pre-trained assumptions, as input tokens in batch mode never attend to output, potentially leading to degraded performance.

% \subsection{Out-Attention Mismatch}
\paragraph{Output-Attention Mismatch}
In the batch-streaming mode, each generated output token $y_k$ can attend to all input tokens $\mathbf{X}$ by KV cache re-encoding:
\begin{equation}
    h_k = \text{CrossAttention}(y_k, \mathbf{X})\quad\quad k\leq j,
\end{equation}
where $y_j$ is the latest generated output token.
However, in interleaved-streaming mode, output tokens can only attend to the subset of inputs $\mathbf{X}_t$ received up to the current step:
\begin{equation}
    h_j^\text{interleaved} = \text{CrossAttention}(y_j, \mathbf{X}_{\leq t}).
\end{equation}
This temporal constraint means that the hidden representation of each decoded token is computed based only on the partial input sequence available at the time, which may lead to inconsistencies compared to batch-mode processing.
% In batch-streaming mode, each output token $y_j$ attends to all input tokens $\mathbf{X}$:
% \begin{equation}
%     h_j = \text{CrossAttention}(y_j, \mathbf{X}).
% \end{equation}
% However, in streaming mode, output tokens can only attend to the subset of inputs received up to the current step:
% \begin{equation}
%     h_j^\text{stream} = \text{CrossAttention}(y_j, \mathbf{X}_{\leq t}).
% \end{equation}
% This temporal constraint means that the hidden representation of each decoded token is computed based only on the partial input sequence available at the time, which may lead to inconsistencies compared to batch-mode processing.

% \subsection{Position-ID Mismatch}
\paragraph{Position-ID Mismatch}
In batch-streaming, tokens receive contiguous position IDs by position re-encoding, so that for an input sequence $\mathbf{X}_{t}$ and output sequence $\mathbf{Y}_{t'}$,
we have:
\begin{equation}
p(x_i) = i, \quad p(y_j) = t + j,
\end{equation}
ensuring that the relative positional differences, \(p(t_j)-p(t_i)\), accurately reflect the true token order and guide the positional embedding function \(g(p(t))\) to generate coherent embeddings. For interleaved-streaming, however, inputs and outputs are interleaved (e.g., \(x_1, y_1, x_2, y_2, \dots\)) while still being assigned continuous IDs from \(1\) to \(n+m\). This misrepresents the true temporal gaps between tokens; the relative differences \(p(t_j)-p(t_i)\) no longer mirror the actual sequence structure. 
\section{Impact Analysis of Mismatches}
Applying batch-trained LLMs to streaming mode introduces structural mismatches.
Existing research has not systematically analyzed the nature of these mismatches between streaming and batch-processing.
We employ a stepwise ablation approach to systematically isolate each mismatch and assess its impact on streaming task performance.

\paragraph{Setup}
This section analyzes the impact of the three mismatches using the streaming text translation task with wait-k reading \& writing policy \cite{ma2019stacl}. 
All experiments are conducted on the IWSLT-17 dataset \cite{cettolo2017overview}, covering two cross-lingual translation tasks: En-Fr and En-De. We use Gemma2-2B-Instruct model \cite{team2024gemma} and Phi3-Mini-Instruct model \cite{abdin2024phi} with 3.8B parameters for all experiments, evaluating model performance using BLEU scores. \cite{post2018call}.

\begin{table*}
\centering
\setlength{\tabcolsep}{7pt}
\small
\renewcommand{\arraystretch}{1.2}

\begin{tabular}{>{\centering\arraybackslash}p{0.9cm} | >{\centering\arraybackslash}p{3.2cm} | >{\raggedright\arraybackslash}p{1.8cm} >{\raggedright\arraybackslash}p{1.8cm} >{\raggedright\arraybackslash}p{1.8cm} >{\raggedright\arraybackslash}p{1.8cm} >{\centering\arraybackslash}p{1.3cm}}
\toprule
\multirow{2}{*}{\textbf{Dataset}} & \multirow{2}{*}{\textbf{Mode}} & \multicolumn{5}{c}{\textbf{Gemma2-2B-Instruct (wait-k)}} \\
\cline{3-7}
& & \multicolumn{1}{c}{1} & \multicolumn{1}{c}{3} & \multicolumn{1}{c}{5} & \multicolumn{1}{c}{7} & \multicolumn{1}{c}{Max. Imp.} \\
\midrule

\multirow{4}{*}{En-Fr} 
& \cellcolor{gray!10} \centering Interleaved-streaming & 
\cellcolor{gray!10} $30.93_{\pm0.08}$  & 
\cellcolor{gray!10} $37.67_{\pm0.11}$  & 
\cellcolor{gray!10} $39.12_{\pm0.09}$  & 
\cellcolor{gray!10} $39.65_{\pm0.07}$  & 
\cellcolor{gray!10} \\
& \cellcolor{white} \centering Batch-streaming (No re.) & 
\cellcolor{white} $33.13_{\pm0.09}$$^{\color{red}{\uparrow 2.20}}$ & 
\cellcolor{white} $39.29_{\pm0.06}$$^{\color{red}{\uparrow 1.62}}$ & 
\cellcolor{white} $40.66_{\pm0.10}$$^{\color{red}{\uparrow 1.54}}$ & 
\cellcolor{white} $40.82_{\pm0.09}$$^{\color{red}{\uparrow 1.17}}$ & 
\cellcolor{white} ${\color{red}{\uparrow 2.20}}$\\
& \cellcolor{gray!10} \centering Batch-streaming (Pos re.) & 
\cellcolor{gray!10}$33.19_{\pm0.07}$$^{\color{red}{\uparrow 0.06}}$ & 
\cellcolor{gray!10}$39.43_{\pm0.13}$$^{\color{red}{\uparrow 0.14}}$ & 
\cellcolor{gray!10}$40.78_{\pm0.08}$$^{\color{red}{\uparrow 0.12}}$ & 
\cellcolor{gray!10}$40.89_{\pm0.07}$$^{\color{red}{\uparrow 0.07}}$ & \cellcolor{gray!10} ${\color{red}{\uparrow 0.14}}$ \\
& \cellcolor{white} \centering Batch-streaming (All re.) & 
\cellcolor{white}$33.47_{\pm0.10}$$^{\color{red}{\uparrow 0.28}}$ & 
\cellcolor{white}$39.62_{\pm0.08}$$^{\color{red}{\uparrow 0.19}}$& 
\cellcolor{white}$40.91_{\pm0.11}$$^{\color{red}{\uparrow 0.13}}$& 
\cellcolor{white}$41.01_{\pm0.09}$$^{\color{red}{\uparrow 0.12}}$& 
\cellcolor{white}${\color{red}{\uparrow 0.28}}$ \\
\midrule

\multirow{4}{*}{En-De} 
& \cellcolor{gray!10} \centering Interleaved-streaming & 
\cellcolor{gray!10} $20.44_{\pm 0.06}$ &
\cellcolor{gray!10} $26.86_{\pm 0.10}$ &
\cellcolor{gray!10} $29.13_{\pm 0.08}$ &
\cellcolor{gray!10} $29.90_{\pm 0.07}$ &
\cellcolor{gray!10} \\
& \cellcolor{white} \centering Batch-streaming (No re.) & 
\cellcolor{white} $21.97_{\pm 0.04}$$^{\color{red}{\uparrow 1.53}}$ &
\cellcolor{white} $28.30_{\pm 0.07}$$^{\color{red}{\uparrow 1.44}}$ &
\cellcolor{white} $30.52_{\pm 0.06}$$^{\color{red}{\uparrow 1.39}}$ &
\cellcolor{white} $31.36_{\pm 0.05}$$^{\color{red}{\uparrow 1.46}}$ &
\cellcolor{white} ${\color{red}{\uparrow 1.53}}$ \\
& \cellcolor{gray!10} \centering Batch-streaming (Pos re.) & 
\cellcolor{gray!10} $22.06_{\pm 0.03}$$^{\color{red}{\uparrow 0.09}}$ & 
\cellcolor{gray!10} $28.38_{\pm 0.05}$$^{\color{red}{\uparrow 0.08}}$ & 
\cellcolor{gray!10} $30.63_{\pm 0.04}$$^{\color{red}{\uparrow 0.11}}$ & 
\cellcolor{gray!10} $31.45_{\pm 0.05}$$^{\color{red}{\uparrow 0.09}}$ & 
\cellcolor{gray!10} ${\color{red}{\uparrow 0.11}}$ \\
& \cellcolor{white} \centering Batch-streaming (All re.) & 
\cellcolor{white} $22.25_{\pm 0.05}$$^{\color{red}{\uparrow 0.19}}$ & 
\cellcolor{white} $28.61_{\pm 0.06}$$^{\color{red}{\uparrow 0.23}}$ & 
\cellcolor{white} $30.77_{\pm 0.07}$$^{\color{red}{\uparrow 0.14}}$ & 
\cellcolor{white} $31.56_{\pm 0.06}$$^{\color{red}{\uparrow 0.11}}$ & 
\cellcolor{white} ${\color{red}{\uparrow 0.23}}$ \\
\midrule

\multirow{2}{*}{\textbf{Dataset}} & \multirow{2}{*}{\textbf{Mode}} & \multicolumn{5}{c}{\textbf{Phi3-Mini-Instruct (wait-k)}} \\
\cline{3-7}
& & \multicolumn{1}{c}{1} & \multicolumn{1}{c}{3} & \multicolumn{1}{c}{5} & \multicolumn{1}{c}{7} & \multicolumn{1}{c}{Max. Imp.} \\
\midrule

\multirow{4}{*}{En-Fr} 
& \cellcolor{gray!10} \centering Interleaved-streaming & 
\cellcolor{gray!10} $29.03_{\pm0.10}$& 
\cellcolor{gray!10} $36.54_{\pm0.14}$& 
\cellcolor{gray!10} $38.42_{\pm0.13}$& 
\cellcolor{gray!10} $39.27_{\pm0.09}$& 
\cellcolor{gray!10} \\
& \cellcolor{white} \centering Batch-streaming (No re.) & 
\cellcolor{white} $30.96_{\pm0.10}$$^{\color{red}{\uparrow 1.93}}$& 
\cellcolor{white} $38.42_{\pm0.08}$$^{\color{red}{\uparrow 1.88}}$& 
\cellcolor{white} $39.80_{\pm0.07}$$^{\color{red}{\uparrow 1.42}}$& 
\cellcolor{white} $40.93_{\pm0.11}$$^{\color{red}{\uparrow 1.66}}$& 
\cellcolor{white} ${\color{red}{\uparrow 1.93}}$\\
& \cellcolor{gray!10} \centering Batch-streaming (Pos re.) & 
\cellcolor{gray!10} $31.08_{\pm0.06}$$^{\color{red}{\uparrow 0.12}}$& 
\cellcolor{gray!10} $38.51_{\pm0.08}$$^{\color{red}{\uparrow 0.09}}$& 
\cellcolor{gray!10} $39.87_{\pm0.12}$$^{\color{red}{\uparrow 0.07}}$& 
\cellcolor{gray!10} $40.96_{\pm0.05}$$^{\color{red}{\uparrow 0.03}}$& 
\cellcolor{gray!10} ${\color{red}{\uparrow 0.12}}$\\
& \cellcolor{white} \centering Batch-streaming (All re.) & 
\cellcolor{white} $31.21_{\pm0.09}$$^{\color{red}{\uparrow 0.20}}$& 
\cellcolor{white} $38.67_{\pm0.13}$$^{\color{red}{\uparrow 0.16}}$& 
\cellcolor{white} $39.98_{\pm0.11}$$^{\color{red}{\uparrow 0.11}}$& 
\cellcolor{white} $41.05_{\pm0.07}$$^{\color{red}{\uparrow 0.09}}$& 
\cellcolor{white} ${\color{red}{\uparrow 0.20}}$\\
\midrule

\multirow{4}{*}{En-De} 
& \cellcolor{gray!10} \centering Interleaved-streaming & 
\cellcolor{gray!10} $20.74_{\pm0.05}$& 
\cellcolor{gray!10} $27.46_{\pm0.14}$& 
\cellcolor{gray!10} $29.56_{\pm0.10}$& 
\cellcolor{gray!10} $30.67_{\pm0.06}$& 
\cellcolor{gray!10} \\
& \cellcolor{white} \centering Batch-streaming (No re.) & 
\cellcolor{white} $22.21_{\pm0.08}$$^{\color{red}{\uparrow 1.47}}$& 
\cellcolor{white} $28.85_{\pm0.11}$$^{\color{red}{\uparrow 1.39}}$& 
\cellcolor{white} $30.88_{\pm0.05}$$^{\color{red}{\uparrow 1.32}}$& 
\cellcolor{white} $31.92_{\pm0.07}$$^{\color{red}{\uparrow 1.25}}$& 
\cellcolor{white} ${\color{red}{\uparrow 1.47}}$\\
& \cellcolor{gray!10} \centering Batch-streaming (Pos re.) & 
\cellcolor{gray!10} $22.28_{\pm0.06}$$^{\color{red}{\uparrow 0.07}}$&
\cellcolor{gray!10} $28.87_{\pm0.09}$$^{\color{red}{\uparrow 0.02}}$&
\cellcolor{gray!10} $30.91_{\pm0.11}$$^{\color{red}{\uparrow 0.03}}$&
\cellcolor{gray!10} $31.95_{\pm0.13}$$^{\color{red}{\uparrow 0.03}}$&
\cellcolor{gray!10} ${\color{red}{\uparrow 0.07}}$ \\
& \cellcolor{white} \centering Batch-streaming (All re.) & 
\cellcolor{white} $22.45_{\pm0.07}$$^{\color{red}{\uparrow 0.17}}$&
\cellcolor{white} $28.98_{\pm0.07}$$^{\color{red}{\uparrow 0.11}}$&
\cellcolor{white} $31.01_{\pm0.07}$$^{\color{red}{\uparrow 0.10}}$&
\cellcolor{white} $32.03_{\pm0.07}$$^{\color{red}{\uparrow 0.08}}$&
\cellcolor{white} ${\color{red}{\uparrow 0.17}}$\\
\bottomrule
\end{tabular}
\caption{The BLEU performance variations reflect the stepwise elimination of mismatches between batch processing and streaming. Interleaved-streaming represents the presence of all three mismatches. Batch-streaming (No re.) corresponds to batch-streaming with interleaved position encoding, where the input-attention mismatch is eliminated. Batch-streaming (Pos re.) further removes the position-ID mismatch through position re-encoding. Finally, Batch-streaming (All re.) eliminates the output-attention mismatch by re-encoding the KV cache. }
\label{tab:Mismatches}
\end{table*}

\paragraph{Effects of Input-Attention Mismatch}
The interleaved-streaming mode, which exhibits all three mismatches, serves as our baseline for comparison. 
The batch-streaming mode eliminates input-attention mismatch by preventing source tokens from attending to generated target tokens. Building on this, we \textit{apply the same positional encoding} as interleaved-streaming within the batch-streaming framework. 
Notably, without KV cache and position embedding re-encoding, the batch-streaming approach still retains both output-attention mismatch and position-ID mismatch. 

Table~\ref{tab:Mismatches} shows that eliminating the input-attention mismatch improves BLEU scores across different wait-k strategies, with a maximum increase of 2.20 on the En-Fr translation task and 1.53 on the En-De translation task. This indicates that processing streaming data in an interleaved streaming manner with a batch-pretrained model leads to performance degradation.

\paragraph{Effects of Position-ID Mismatch}

Re-encoding can address the remaining two mismatches. We further decompose re-encoding into two components: KV cache re-encoding and position embedding re-encoding. The former enables target tokens to attend to the most recently available tokens, thereby resolving the output-attention mismatch. The latter corrects the position-ID mismatch by adjusting position embeddings to align with the streaming paradigm. 
Expanding on this, the batch-streaming paradigm with position embedding re-encoding further resolves the position-ID mismatch while still retaining output-attention mismatch.

Table~\ref{tab:Mismatches} shows that position embeddings re-encoding does not lead to significant performance improvements, with a maximum gain of only 0.14 on the En-Fr and En-De translation tasks. This suggests that position-ID mismatch is not a primary factor affecting streaming task performance, challenging previous claims regarding the role of positional encoding in streaming models.

\paragraph{Effects of Output-Attention Mismatch}
The KV cache re-encoding can address the remaining output-attention mismatch. On the setting of the former, incorporating both KV cache and position embedding re-encoding into batch-streaming paradigm eliminates all mismatches, making it closely resemble the batch-processing setting.

Table~\ref{tab:Mismatches} shows that re-encoding previously generated tokens also does not significantly improve model performance. Although re-encoding allows target tokens to attend to the most recent input context, the inherent constraints of streaming tasks prevent already generated outputs from being modified. As a result, re-encoding does not alter the fundamental partial information nature of streaming tasks; instead, its primary effect is to correct the generation path of subsequent tokens. However, experimental results indicate that this correction is not a decisive factor in performance improvement.

\begin{table*}[t]
    \centering
    \footnotesize
    \setlength{\tabcolsep}{7pt}
    \renewcommand{\arraystretch}{1.2} 
    \begin{tabular}{l|c|cccc|cccc}
        \toprule
         \multirow{2}{*}{Model} & \multirow{2}{*}{Position setting} & \multicolumn{4}{c}{En-Fr (Wait-k)} & \multicolumn{4}{c}{En-De (Wait-k)} \\
        \cmidrule(lr){3-6} \cmidrule(lr){7-10}
        & &  1 & 3 & 5 & 7 & 1 & 3 & 5 & 7 \\
        \midrule
        \multirow{4}{*}{\rotatebox{0}{Gemma2-2B-Instruct}}
         & \cellcolor{gray!10}Remove all pos.  & 
         \cellcolor{gray!10}27.11 & \cellcolor{gray!10}34.98 & \cellcolor{gray!10}37.54 & \cellcolor{gray!10}38.02 & 
         \cellcolor{gray!10}19.01 & \cellcolor{gray!10}25.93 & \cellcolor{gray!10}27.71 & \cellcolor{gray!10}28.87  \\
         & Remove source pos.  & 28.35 & 36.12 & 38.42 & 39.03 & 19.63 & 26.82 & 28.08 & 29.36 \\
         & \cellcolor{gray!10}Remove target pos.  & 
          \cellcolor{gray!10}29.14 & \cellcolor{gray!10}36.83 & \cellcolor{gray!10}39.01 & \cellcolor{gray!10}39.62 & 
          \cellcolor{gray!10}19.91 & \cellcolor{gray!10}27.01 & \cellcolor{gray!10}28.59 & \cellcolor{gray!10}29.51 \\
         & Retain all pos. & \textbf{33.23} & \textbf{39.39} & \textbf{40.76}& \textbf{40.92} & \textbf{22.35} & \textbf{28.88} & \textbf{30.84} & \textbf{31.47} \\
        \midrule
        \multirow{4}{*}{\rotatebox{0}{\shortstack{Phi3-Mini-Instruct \\ (3.8B)}}}
         & \cellcolor{gray!10}Remove all pos.  & 
         \cellcolor{gray!10}26.73 & \cellcolor{gray!10}34.85 & \cellcolor{gray!10}37.31 & \cellcolor{gray!10}37.92 & 
         \cellcolor{gray!10}18.86 & \cellcolor{gray!10}25.87 & \cellcolor{gray!10}27.79 & \cellcolor{gray!10}29.01 \\
         & Remove source pos.  & 27.98 & 35.96 & 38.17 & 38.95 & 19.47 & 26.78 & 28.19 & 29.54 \\
         & \cellcolor{gray!10}Remove target pos.  & 
         \cellcolor{gray!10}28.84 & \cellcolor{gray!10}36.58 & \cellcolor{gray!10}39.04 & \cellcolor{gray!10}39.46 &  
         \cellcolor{gray!10}19.83 & \cellcolor{gray!10}26.95 & \cellcolor{gray!10}28.64 & \cellcolor{gray!10}29.78 \\
         & Retain all pos. &\textbf{30.96} &\textbf{38.45}  & \textbf{39.89} & \textbf{40.57} & \textbf{22.21} & \textbf{28.86} & \textbf{30.92} & \textbf{31.94} \\
        \bottomrule
    \end{tabular}
    \caption{Effect of source and target position removal on streaming LLMs performance. We simulate position removal by assigning a constant position ID of 0 to all tokens instead of removing the positional embeddings.}
    \label{tab:remove_pos}
\end{table*}

Our experiments demonstrate that input-attention mismatch significantly impacts streaming translation performance, highlighting the performance gains of using a batch-processing architecture for streaming tasks.\footnote{We provide the detailed training process for different settings in the Appendix \ref{Appendix-B3}.} 
On the other hand, contrary to existing studies \cite{raffel-etal-2024-simultaneous,guo-etal-2024-decoder,he2024let}, position-ID mismatch is not the primary reason for re-encoding, and interleaved positional encoding achieves performance comparable to continuous position encoding in batch processing. 
To investigate the discrepancy between our findings with the common assumption, we conduct a comprehensive analysis of \textbf{how position encoding impacts LLMs in streaming scenarios.}

\section{Impact Analysis of Position Encoding}

The above analysis suggests that positional mismatches do not significantly impact the performance of streaming tasks.
To further elucidate this phenomenon, this section provides a detailed investigation into the impact of positional encoding on the performance of LLMs in streaming scenarios.

\begin{table*}[t]
    \centering
    \footnotesize
    \setlength{\tabcolsep}{3.5pt}
    \renewcommand{\arraystretch}{1.2} 
    \begin{tabular}{l|c|cccccc|cccccc}
        \toprule
         \multirow{2}{*}{Model} & \multirow{2}{*}{Wait-k} & \multicolumn{6}{c}{En-Fr (Target start id $\phi$)} & \multicolumn{6}{c}{En-De (Target start id $\phi$)} \\
        \cmidrule(lr){3-8} \cmidrule(lr){9-14}
        & &  0 & 0.5 & 128 & 256 & 512 & $\Delta$ & 0 & 0.5 & 128 & 256 & 512 & $\Delta$\\
        \midrule
        \multirow{4}{*}{\rotatebox{0}{Gemma2-2B-Instruct}}
         & \cellcolor{gray!10}5  & \cellcolor{gray!10}40.76 & \cellcolor{gray!10}40.76 & \cellcolor{gray!10}40.70 & \cellcolor{gray!10}40.57 & \cellcolor{gray!10}40.68 & \cellcolor{gray!10}\textit{0.19} & \cellcolor{gray!10}30.84 & \cellcolor{gray!10}30.84 & \cellcolor{gray!10}30.90 & \cellcolor{gray!10}30.80 & \cellcolor{gray!10}30.95 & \cellcolor{gray!10}\textit{0.15 }\\
         & 7  & 40.92 & 40.92 & 40.85 & 40.91 & 40.92 & \textit{0.07} & 31.47 & 31.47 & 31.44 & 31.57 & 31.67 & \underline{\textit{0.23}} \\
         & \cellcolor{gray!10}9  & \cellcolor{gray!10}40.91 & \cellcolor{gray!10}40.91 & \cellcolor{gray!10}40.90 & \cellcolor{gray!10}40.88 & \cellcolor{gray!10}40.97 & \cellcolor{gray!10}\textit{0.09} & \cellcolor{gray!10}31.73 & \cellcolor{gray!10}31.73 & \cellcolor{gray!10}31.87 & \cellcolor{gray!10}31.91 & \cellcolor{gray!10}31.88 & \cellcolor{gray!10}\textit{0.18} \\
         & 11 & 41.10 & 41.10 & 41.14 & 40.96 & 41.05 & \textit{0.18} & 31.95 & 31.95 & 31.98 & 31.95 & 31.89 & \textit{0.09} \\
        \midrule
        \multirow{4}{*}{\rotatebox{0}{\shortstack{Phi3-Mini-Instruct \\ (3.8B)}}}
        & \cellcolor{gray!10}5  & \cellcolor{gray!10}39.89 & \cellcolor{gray!10}39.89 & \cellcolor{gray!10}39.91 & \cellcolor{gray!10}40.06 & \cellcolor{gray!10}39.87 & \cellcolor{gray!10}\textit{0.19} & \cellcolor{gray!10}30.92 & \cellcolor{gray!10}30.92 & \cellcolor{gray!10}30.76 & \cellcolor{gray!10}30.81 & \cellcolor{gray!10}30.86 & \cellcolor{gray!10}\textit{0.16} \\
        & 7  & 40.57 & 40.57 & 40.53 & 40.72 & 40.71 & \textit{0.19} & 31.94 & 31.94 & 31.78 & 31.84 & 31.78 & \textit{0.16} \\
        & \cellcolor{gray!10}9  & \cellcolor{gray!10}41.31 & \cellcolor{gray!10}41.31 & \cellcolor{gray!10}41.24 & \cellcolor{gray!10}41.35 & \cellcolor{gray!10}41.44 & \cellcolor{gray!10}\textit{0.20} & \cellcolor{gray!10}32.18 & \cellcolor{gray!10}32.18 & \cellcolor{gray!10}32.10 & \cellcolor{gray!10}32.21 & \cellcolor{gray!10}32.09 & \cellcolor{gray!10}\textit{0.12} \\
        & 11 & 41.92 & 41.92 & 42.03 & 41.94 & 41.93 & \textit{0.11} & 32.26 & 32.26 & 32.23 & 32.33 & 32.28 & \textit{0.10} \\
        \midrule
        \multirow{4}{*}{\rotatebox{0}{LLama3.1-8B-Instruct}}
        & \cellcolor{gray!10}5  & \cellcolor{gray!10}40.11 & \cellcolor{gray!10}40.11 & \cellcolor{gray!10}40.10 & \cellcolor{gray!10}39.93 & \cellcolor{gray!10}39.92 & \cellcolor{gray!10}\textit{0.19} & \cellcolor{gray!10}30.33 & \cellcolor{gray!10}30.33 & \cellcolor{gray!10}30.21 & \cellcolor{gray!10}30.37 & \cellcolor{gray!10}30.34 & \cellcolor{gray!10}\textit{0.16} \\
        & 7  & 40.30 & 40.30 & 40.32 & 40.35 & 40.31 & \textbf{\textit{0.03}} & 31.23 & 31.23 & 31.18 & 31.16 & 31.25 & \textit{0.09} \\
        & \cellcolor{gray!10}9  & \cellcolor{gray!10}40.15 & \cellcolor{gray!10}40.15 & \cellcolor{gray!10}40.32 & \cellcolor{gray!10}40.34 &\cellcolor{gray!10} 40.35 & \cellcolor{gray!10}\textit{0.20} & \cellcolor{gray!10}31.80 & \cellcolor{gray!10}31.80 &\cellcolor{gray!10} 31.83 & \cellcolor{gray!10}31.76 & \cellcolor{gray!10}31.89 & \cellcolor{gray!10}\textit{0.13} \\
        & 11 & 40.53 & 40.53 & 40.47 & 40.58 & 40.63 & \textit{0.16} & 32.04 & 32.04 & 31.98 & 32.07 & 32.08 & \textit{0.10} \\
        \bottomrule
    \end{tabular}
    \caption{Performance comparison of different models with various wait-k policies and target start IDs. $\Delta$ represents the range of variation in BLEU scores when the start id of target token takes different values. We use bold to indicate the smallest variation and underline to represent the largest variation.}
    \label{tab:wait-k-text}
\end{table*}

\subsection{Is Position Encoding Necessary for Streaming Tasks?}
% \paragraph{Setup}

Building upon the experimental setup from the previous section, we further investigate the necessity of positional encoding in streaming tasks by separately removing global positional encoding and target-side positional encoding. 
Table~\ref{tab:remove_pos} presents the BLEU scores on the En-Fr and En-De streaming translation tasks  after removing position encodings at different locations. We simulate position removal by assigning a constant position ID of 0 to all tokens instead of removing the positional encoding module. For the setting of position-retaining, we apply interleaved positional encoding as illustrated in previous section.
The table reveals that removing positional information from either the source or target side results in a clear performance degradation, with the maximum drop exceeding 10\%.
In contrast, when both source and target positional information are removed, the model maintains roughly 80\% of its BLEU score compared to the fully position-retaining setting.

This finding aligns with previous studies suggesting that LLMs can still learn certain positional information even without explicit positional encoding \cite{haviv2022transformer}. However, it is important to emphasize that positional encoding remains relevant for streaming tasks, particularly on the target side. Notably, the absence of target-side positional encoding leads to a measurable performance decline, highlighting its role in maintaining effective token generation in streaming scenarios.

\subsection{Group Position Encoding Is An Option for Streaming Tasks}
\begin{figure}[t]
    \centering
    \includegraphics[width=\linewidth]{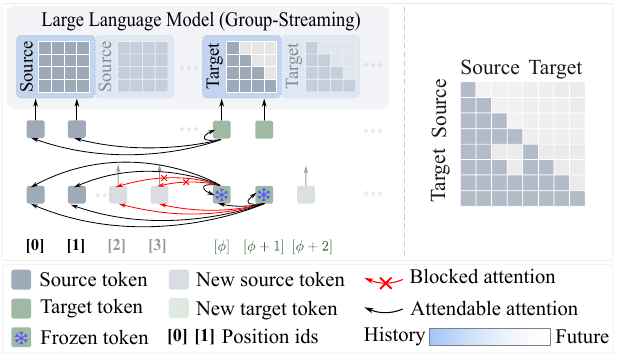}
    \caption{Framework of our \textit{Group-streaming} LLMs. (Left) Positional grouping of source and target tokens in the streaming LLM, avoiding re-encoding. The group start ID $\phi$ is a hyperparameter. (Right) The attention mask matrix during the training ensures that target tokens can only attend to locally available inputs.}
    \label{Fig:Model}
\end{figure}
Given that positional encoding is necessary and interleaved positional encoding has minimal impact on streaming task performance, one might question whether streaming problems can be modeled using interleaved positional encoding and batch-streaming mode. However, this is not an optimal choice, as \textbf{interleaved positional encoding lacks direct generalizability to batch processing.}

In real-world scenarios, the target sequence is not available in advance, making it impossible to predefine source positions. This limitation hinders the generalization of streaming models to offline settings. To address this issue, we propose \textbf{a group position encoding} based on batch-streaming framework for streaming LLMs as shown in Figure~\ref{Fig:Model}, where source and target tokens are independently assigned positional encodings, ensuring only monotonic continuity within each group. 
Specifically, in our proposed approach, the source position encoding remains consistent with batch processing mode, starting from 0. In contrast, the target position begins from a predefined starting value $\phi$.

This approach makes it feasible to prefill source position encodings even without target information and naturally extends to batch processing.
In fact, interleaved position encoding can be viewed as a special case of group position encoding, with the distinction that the interleaved mode uses non-uniform positional intervals.

\subsection{What is the Impact of Group Position Offset on Model Performance?}

This section provides a detailed discussion on the impact of target position offset $\phi$.
% on streaming tasks.

\paragraph{Setup}
We evaluate the impact of grouped positional encoding on text translation and automatic speech recognition tasks. For text translation, we use the IWSLT-17 dataset, focusing on En-Fr and En-De translation tasks, with models including Gemma2-2B-Instruct \cite{team2024gemma}, Phi3-Mini-Instruct \cite{abdin2024phi}, and LLaMA3.1-8B-Instruct \cite{dubey2024llama}. For ASR, we use the LibriSpeech dataset \cite{panayotov2015librispeech}, with Phi3 as the selected model.
Translation performance is assessed using BLEU scores, while ASR performance is evaluated based on WER \cite{radford2023robust}.
Detailed experimental settings and hyperparameters are provided in the appendix.

\paragraph{Results}
The streaming text translation task results in Table~\ref{tab:wait-k-text} and streaming ASR task in Table~\ref{tab:ASR} indicate that varying the initial offset of the target-side group position encoding within a reasonable range does not significantly affect the performance of LLMs in streaming scenarios. This suggests that the model is highly robust to the choice of initial group position offset.
Specifically, when the offset is set to 0, the source and target positions are fully overlap, whereas an offset of 0.5 results in complete separation. Despite this contrast, both settings yield comparable performance, suggesting that positional overlap appears to have limited impact on the effectiveness of group position encoding.

\begin{table}[tbp]
    \centering
    \footnotesize
    \setlength{\tabcolsep}{7pt}
    \renewcommand{\arraystretch}{1.2} 
    \begin{tabular}{c|ccccccc}
        \toprule
         \multirow{2}{*}{Wait-k} & \multicolumn{6}{c}{Speech-Text (Target start id $\phi$)} \\
        \cmidrule(lr){2-7} 
        &0&  256 & 512 & 1024 & 2048 & $\Delta$ \\
        \midrule
        1 & 6.02 & 6.05 & 6.04 & 6.07 & 6.17 & \underline{\textit{0.15}}  \\
        3 & 4.12 & 4.10 & 4.09 & 4.08 & 4.19  & {\textit{0.11}}  \\
        5 & 3.52 & 3.58 & 3.55 & 3.59 & 3.61 & \textbf{\textit{0.09}}  \\
        7 & 3.33 & 3.33 & 3.38 & 3.41 & 3.45 & \textit{0.12}  \\
        \bottomrule
    \end{tabular}
    \caption{Performance of Phi3 with various wait-k policies and target start IDs. $\Delta$ represents the range of variation in WER scores when the start id of target token takes different values. }
    \label{tab:ASR}
\end{table}

\subsection{Why Group Position Encoding Works?}
The RoPE encodes relative position information via rotation matrices $R$ applied to each token's query and key: $q^r_n = R(n)q_n$ and $k^r_n = R(n)k_n$, where $n$ denotes the position ID. Then the dot product attention score can be written as $Attn(n,\text{cache})= \sum_{i} {q_{n}^{r}}^T{k_i}^r= \sum_{i}q_n^T R^T(n)R(i)k_i=\sum_{i=0}^{S+n} q_n^T R(n - i) k_i$, where $S$ is the token length of source input and $q_n$ is query of a target token. The relative position can be written as $\Delta = n-i = \phi + j - i$, where $j$ denotes the index of the target token and $\phi$ represents the position offset between the first token of the target and that of the source. We split the above dot-product attention into two parts: target-to-target and target-to-source computations: 
\vspace{-0.5\baselineskip}
\begin{align}
    Attn(n,\text{cache})=&\sum_{i=0}^{j} q_n^T R(j - i) k_i \notag \\
    +&\sum_{i=0}^{S} q_n^T R(\phi+j - i) k_i.
\end{align}
The relative position $j-i$ in the first target-to-target term remains consistent across both RoPE and group position encoding. For cross-segment attention in target-to-source, the difference of relative position between RoPE and group position encoding is determined by the position offset $\phi$. In original RoPE, $\phi$ equals the length of the source sequence and varies with input length, whereas in group position encoding, $\phi$ is predefined as a fixed constant.
LLMs are capable of easily learning and internalizing the semantics of the relative offset by fine-tuning. Once the model has correctly understood the meaning of $\phi$ as a position shift, it can accurately capture and assign position relationships across segments, without requiring explicit differentiation between source and target token IDs.\footnote{The detailed analysis can be found in Appendix \ref{Appendix-E}.}

LLMs can learn the position offset $\phi$ through simple fine-tuning, so typical values of $\phi$ do not significantly impact performance. However, when $\phi$ becomes extremely large, it may lead to discrepancies with the model’s pretraining distribution due to the limited context length used during pretraining. 
% Therefore, a reasonable range for $\phi$ should ensure that the maximum relative distance, between the last target token and the first source token, does not exceed the model’s pretraining context length.
Therefore, a reasonable range for $\phi$ should ensure that the maximum relative distance between the last target token and the first source token remains within the model’s pretraining context length.\footnote{We provide additional experiments to demonstrate the potential edge in Appendix \ref{Appendix-D3}.}

% This ensures that relative position gaps remain closer to those encountered during pretraining, thereby facilitating faster convergence and improved performance.
We recommend using a relatively small $\phi$, ideally below the input sentence length, to keep relative position gaps closer to the pretraining distribution, which may facilitate faster convergence and better performance. Notably, when $\phi = 0$, the target starting token is positioned closer to the source starting token and farther from the source ending token. This configuration better reflects the sequential input arrival pattern in streaming scenarios, leading to more stable learning dynamics and enhanced model alignment.

\subsection{Visualization of Streaming Attention}
Taking text translation as an example, we visualize the extent to which each target token attends to past source information during inference. Notably, we normalize the attention weights column-wise (i.e., across each source token) to the range [0, 1]. This normalization offers two key benefits:
(1) it mitigates the influence of tokens with inherently large absolute attention values and highlights the relative importance of attention distribution, making attention strength more interpretable; and
(2) it provides a clearer view of \textit{how each source token distributes its attention across different target tokens.}

\begin{figure}[t]
    \centering
    \includegraphics[width=1\linewidth]{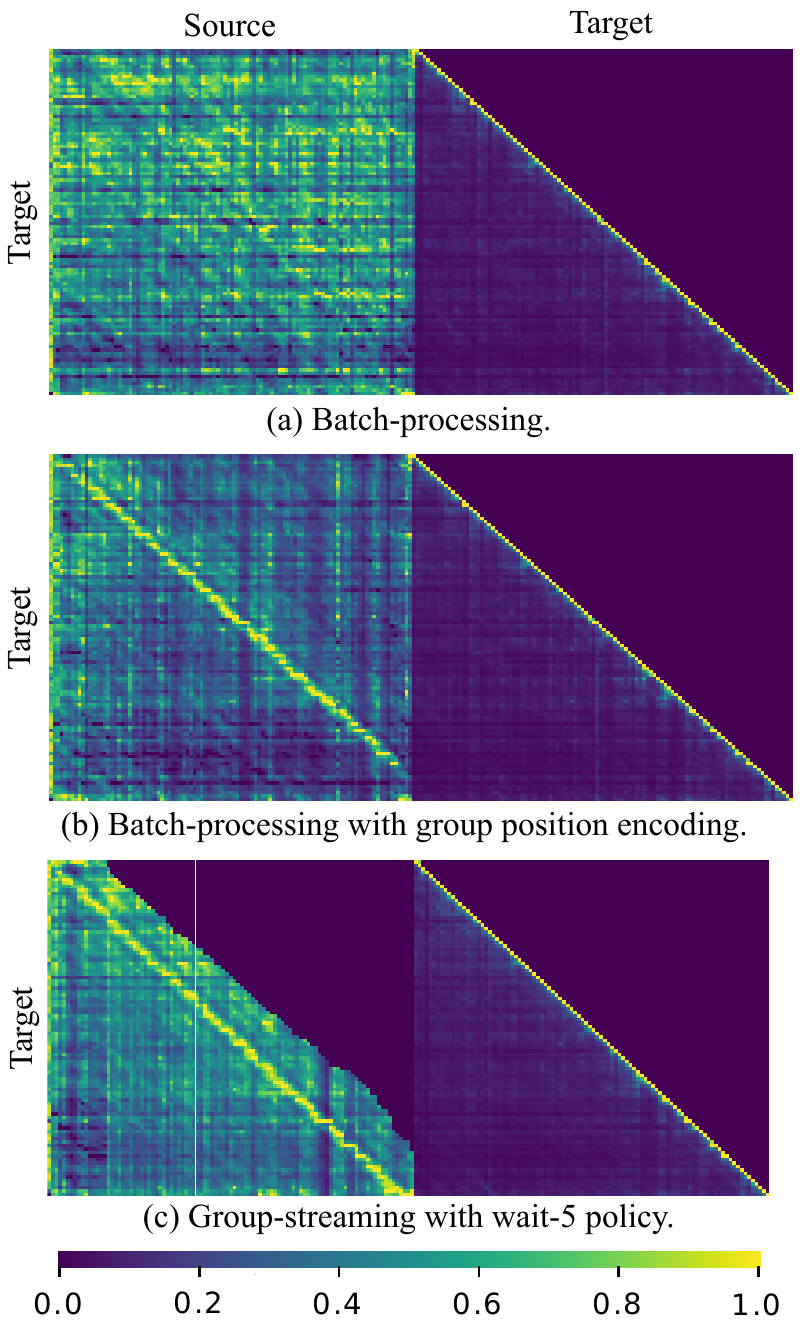}
    \caption{An example of the attention distribution of target tokens, where the attention values of each target token are normalized to emphasize the relative focus. The sample is from IWSLT-17 En-Fr dataset. \protect\footnotemark}
    \label{Fig:Attnmap}
\end{figure}

As shown in Figure~\ref{Fig:Attnmap}, under the batch setting, source tokens distribute their attention uniformly across all target tokens, reflecting a globally constrained behavior. In other words, each target token tends to attend equally to the same source token.
In contrast, with group position encoding, source tokens tend to assign more attention to target tokens with similar positional indices. That is, source tokens are less likely to attend to future target tokens. This observation supports our earlier finding that re-encoding previously generated target tokens offers limited performance gain in streaming tasks under group position encoding.

Moreover, the results in Figure~\ref{Fig:Attnmap} indicate that employing group position encoding in the batch-processing setting shifts the target tokens' attention to the source context along the diagonal direction when the offset $\phi = 0$. This adjustment encourages target tokens to focus more on the currently available input, making the model's behavior more aligned with the requirements of streaming tasks.

% \section{Analysis}
% \label{Analysis}

\section{Discussion}
\label{Discussion}
% \subsection{Why LLMs?}
\paragraph{Why LLMs?}

\footnotetext{Note that the attention values have been normalized. The values do not represent the actual magnitude of attention.}

\begin{figure*}[!t]
    \centering
    \includegraphics[width=\linewidth]{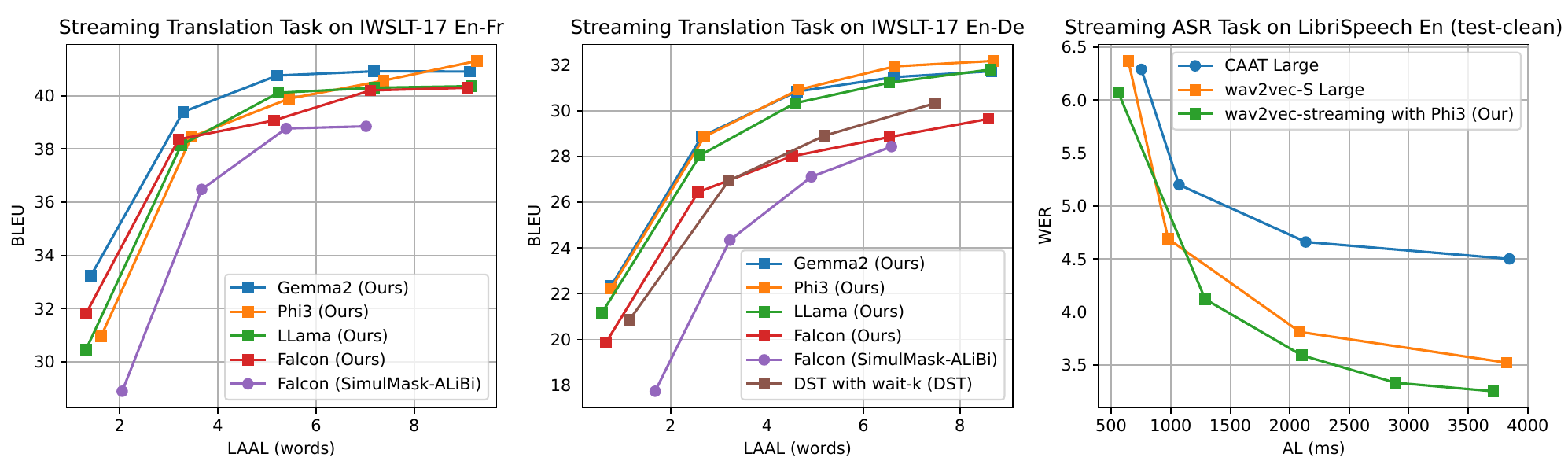}
    \caption{The performance comparison between group position streaming LLMs with other decoder-only models.}
    \label{Fig:Why}
\end{figure*}

% This part will present results in comparison with the SOTA decoder-only methods, highlighting the advantages of using LLMs in streaming tasks.
We apply the proposed group-streaming approach to mainstream large language models and compare its performance against other decoder-only streaming models to highlight its advantages. To demonstrate the effectiveness of our method, we evaluate it on the En-Fr and En-De translation tasks from the IWSLT-17 dataset, as well as the ASR task from the Librispeech dataset. The baselines for text translation include SimulMask \cite{raffel-etal-2024-simultaneous} and DST \cite{guo-etal-2024-decoder}, while the baselines for ASR include CAAT \cite{liu-etal-2021-cross} and Wav2Vec-S \cite{fu2024wav2vec}.
% \paragraph{Text Translation} 
% % For streaming text translation task, we choose the SimulMask and DST as baseline, 

% \paragraph{Automatic Speech Recognition}
% % \paragraph{Speech Translation}

% \subsection{Efficiency Analysis}
% % This part will analyze the complexity of re-encoding and our group-based position encoding.
% The results show that, compared to other decoder-only methods, our group positional encoding-based streaming LLM achieves the best performance in both text translation and speech recognition streaming tasks.

As shown in Figure~\ref{Fig:Why}, the vertical axis represents task-specific performance metrics—BLEU for translation and WER for ASR—while the horizontal axis indicates the model's average latency (AL and LAAL), measured by the number of waited words in translation and the waiting time in ASR. The results show that group-streaming LLMs consistently outperform specialized decoder-only baselines, typically achieving higher accuracy under the same latency conditions.

% \subsection{Generalization}
\paragraph{Generalization}
% \paragraph{Streaming Models in Offline Tasks} 
% \paragraph{Grouped Position Encoding Across Domains}
% We generalize our group position encoding to batch-processing,
% % \paragraph{Intra-Language Generalization} 

\begin{figure}[t]
    \centering
    \includegraphics[width=\linewidth]{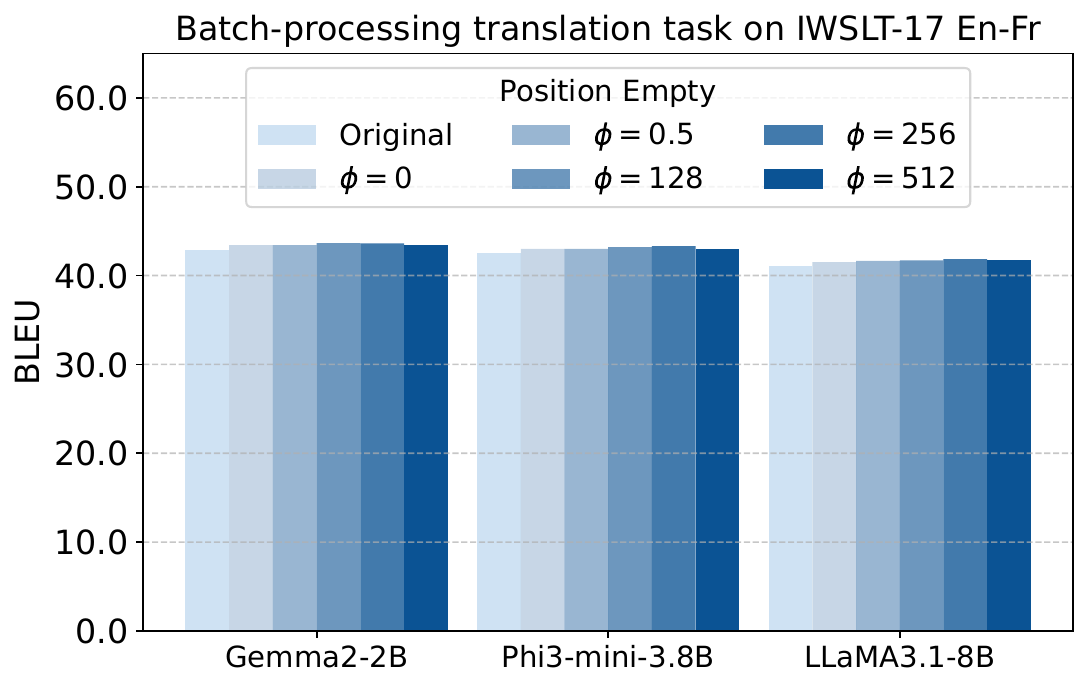}
    \caption{The BLEU performance of batch-processing translation task on IWSLT-17 En-Fr dataset.}
    \label{Fig:bar}
\end{figure}

% We extend our group position encoding to batch processing, and the results in Figure~\ref{Fig:bar} indicate that it does not degrade batch-processing model performance. This demonstrates the strong cross-domain generalization capability of our approach.
We extend our group position encoding to batch processing. The first bar in Figure~\ref{Fig:bar} represents the model that is specifically trained for batch processing using the original RoPE. Subsequently, we applied group position encoding to the batch processing scenario and fine-tuned the model.
The results demonstrate that applying group position encoding introduces no performance degradation for batch processing, confirming its compatibility and generalization across both streaming and batch processing settings.

\section{Related Work}
\label{RelatedWork}
\paragraph{Streaming Language and Speech Transformers}
A typical implementation of Transformer-based streaming  tasks adopts an incremental encoding strategy on the encoder side and an incremental decoding strategy on the decoder side \cite{ma2021streaming, ma2023non,zhang2023hidden}. With the rise of large language models, researchers have begun exploring how to adapt decoder-only architectures for streaming tasks. Among these approaches, batch-streaming models based on prompt structures attempt to approximate offline batch processing by re-encoding tokens during streaming inference \cite{agostinelli-etal-2024-simul,guo2024sillm,koshkin-etal-2024-transllama,wang-etal-2024-simultaneous}.
Some studies suggest that position confusion in streaming environments is a key factor necessitating re-encoding in LLMs \cite{guo-etal-2024-decoder,raffel-etal-2024-simultaneous}. To address this issue, one line of research focuses on modifying the decoder-only architecture to enhance its adaptability to streaming tasks \cite{guo-etal-2024-decoder,tsunoo2024decoder,chen2024streaming}, while another emphasizes optimizing positional encoding—such as the ALIBI positional encoding—to mitigate the effects of incremental position shifts during streaming decoding \cite{raffel-etal-2024-simultaneous}.
In contrast to simulated batch processing, an interleaved-streaming paradigm \cite{du2024cosyvoice,yang2024interleaved} that adheres to temporal order has been explored, wherein input and output tokens are interleaved and encoded sequentially.
While significant progress has been made in developing streaming models, existing studies lack a rigorous analysis of the fundamental differences between batch processing and streaming paradigms.

\paragraph{Position Encoding in Transformers}
Position encoding \cite{raffel2020exploring,presstrain,su2024roformer} is a crucial component of Transformer models \cite{vaswani2017attention}, designed to break the permutation-invariant nature of self-attention mechanisms.
Recent studies have demonstrated that decoder-only Transformer models can still capture positional information even in the absence of explicit positional encoding \cite{shen2018disan}. A plausible explanation is that causal attention masks enforce position-dependent token interactions, implicitly encoding positional information \cite{haviv2022transformer,tsai2019transformer}. Related research has shown that in tasks such as speech modeling \cite{likhomanenko2021cape} and language modeling \cite{haviv2022transformer}, decoder-only Transformers without positional encoding can achieve performance comparable to standard decoder-based Transformers. Furthermore, other studies \cite{kazemnejad2024impact, ruoss2023randomized} have indicated that the generalization ability of Transformers without positional encoding does not degrade significantly when handling varying context lengths.
While significant progress has been made in understanding positional encoding in LLMs, existing research has primarily focused on static scenarios. In contrast, the role of positional encoding in streaming scenarios remains underexplored, where the dynamic modeling of positional information may follow different patterns and exert distinct effects.

\section{Conclusion}
\label{Conclusion}
% This work provides a systematic analysis of the core mismatches between batch-trained LLMs and streaming tasks, revealing that input-attention mismatch is the primary bottleneck, while output-attention and position-ID mismatches have negligible impact. Challenging the prevailing assumption that position re-encoding is necessary, our findings demonstrate its limited utility and highlight the importance of preserving relative token order.
% Based on the findings, we proposed group streaming paradigm, a simple yet effective strategy that bridges the gap between streaming and batch modes without requiring re-encoding. The proposed approach is generalizable and model-agnostic, achieving strong performance across both cross-lingual and cross-modal streaming tasks.

This work provides a systematic analysis of the mismatches that arise in adapting batch-trained LLMs to streaming tasks. We identify input-attention mismatch as the primary bottleneck, while output-attention and position-ID mismatches have negligible impact, challenging the prevailing assumption that position inconsistencies necessitate frequent re-encoding. To clarify this, we conduct the first in-depth analysis of position encoding in streaming settings, showing that preserving strict absolute positions is unnecessary; instead, maintaining relative token order within source and target contexts is more critical.
Building on the insights, we propose the\textit{ group streaming paradigm}, a simple yet effective strategy that bridges the gap between streaming and batch modes without requiring re-encoding. The approach is model-agnostic and generalizable, achieving strong performance across both cross-lingual and cross-modal streaming tasks.

\section*{Limitations}
\label{Limitations}

This paper primarily focuses on exploring the optimal paradigm for streaming models and, therefore, does not delve into different waiting policies. The conclusions drawn in this study have only been validated under the wait-k policy.  Additionally, our study is confined to streaming tasks in the text and audio modalities, with video streaming left for future investigation.

\section*{Acknowledgements}
We thank EIT and IDT High Performance Computing Center for providing computational resources for this project. This work was supported by the  2035 Key Research and Development Program of Ningbo City under Grant No.2024Z127.

% % Bibliography entries for the entire Anthology, followed by custom entries
% %\bibliography{anthology,custom}
% % Custom bibliography entries only
% \bibliography{custom}
% \bibliographystyle{abbrv}
\bibliography{reference}

%% Appendix
% \twocolumn[\newpage]
\setcounter{figure}{0}   
\setcounter{table}{0}
\setcounter{equation}{0}

\appendix
\onecolumn
\section{Different Paradigms on Streaming Tasks}
\begin{figure}[ht]
    \centering
    \includegraphics[width=0.5\linewidth]{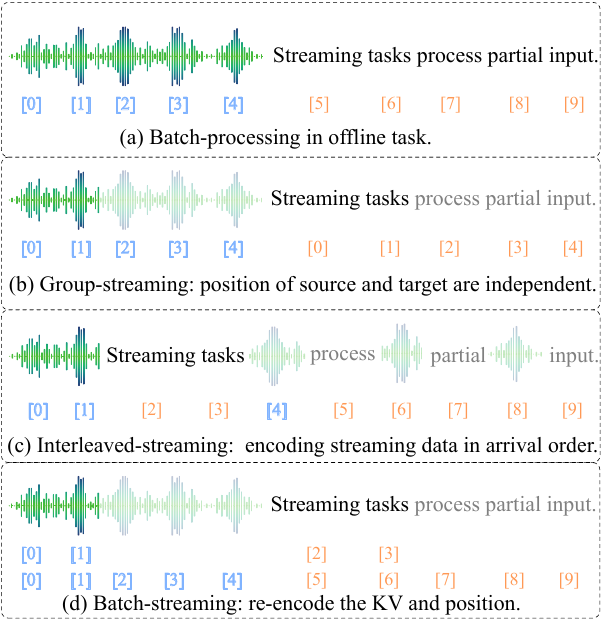}
    \caption{An ASR example for illustration of different paradigms for LLMs processing.}
    \label{Fig:Appendix-intro}
\end{figure}

The main text introduces three approaches for applying LLMs to streaming tasks: batch-streaming, interleaved-streaming, and group-streaming. Among them, batch-streaming maximally simulates the batch-processing paradigm in offline scenarios through re-encoding, with the only difference being the availability of local information in a streaming setting. Figure~\ref{Fig:Appendix-intro} illustrates different paradigms of LLM data processing using ASR as an example. 

We clarify that re-encoding refers to reprocessing all previously generated target tokens after each new source context is read, before generating the next target token. This is solely for optimizing the generation of the latest token without altering previously output content. We exclude scenarios where re-encoding continuously adjusts already output content, as the final alignment after reading the entire input would be equivalent to batch processing. In this context, re-encoding clearly holds positive value.

\section{Model Details}
\label{Model-Details}

\subsection{Model Structure}
\paragraph{Streaming Text LLM}
The group-streaming model, as previously introduced, is designed to enforce a strict attention constraint where historically generated tokens are prevented from attending to newly received source tokens, ensuring a clear separation between past and present information. Additionally, the model maintains independent positional encoding for both source and target tokens, preserving structural integrity while facilitating effective streaming processing.

\paragraph{Streaming Speech LLM}

\begin{figure*}[ht]
    \centering
    \includegraphics[width=0.6\linewidth]{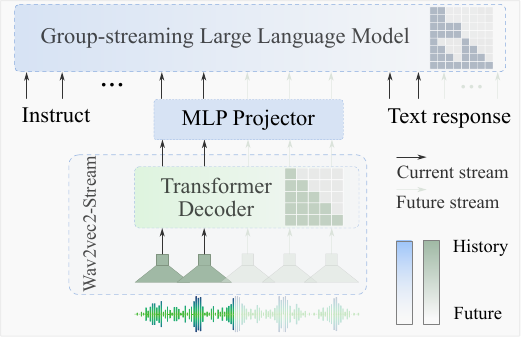}
    \caption{Illustration of our Group-streaming speech Large Language Model, where the group-streaming LLM and the streaming audio encoder are connected through an MLP projector.}
    \label{Fig:Appendix-model}
\end{figure*}

Figure~\ref{Fig:Appendix-model} illustrates the structure of the streaming ASR model proposed in this paper. The model consists of a streaming audio encoder, an MLP projector, and our Group Positional Encoding-based Streaming LLM.

The streaming audio encoder is a variant of Wav2vec2 \cite{baevski2020wav2vec}, with the following key modifications: (1) Positional Encoding Adjustment: We replace the convPE in Wav2vec2 with a causal convolution-based positional encoding (causal ConvPE) to enforce directional constraints on the information flow. (2) Structural Optimization: The Transformer Encoder in Wav2vec2 is replaced with a Transformer Decoder to ensure global unidirectional information constraints, thereby enhancing incremental encoding capability. We refer to this modified model as Wav2vec2-Streaming. While it shares some similarities with Wav2vec-S \cite{fu2024wav2vec}, the latter employs absolute sinusoidal positional encoding, whereas Wav2vec2-Stream retains causal convolution to improve temporal modeling. Additionally, we have modified Wav2vec2 within the HuggingFace Transformers framework\footnote{https://huggingface.co/facebook/wav2vec2-large-960h-lv60-self} to enable seamless interoperability with existing LLMs.  Our code and pretrained weights will be open-sourced for research and application purposes.

Wav2vec2-Stream processes audio data sampled at 16 kHz, where each segment consists of 400 samples, with an 80-sample overlap between consecutive segments. This results in an embedding vector for the LLM approximately every 20 ms, ensuring a fine-grained temporal resolution for streaming speech recognition.
Similar to \cite{chen2021direct,dong2022learning} et al., we adopt a fixed-interval audio segmentation approach combined with the wait-k strategy for streaming tasks. In our setup, the time interval is set to 400 ms, ensuring a structured and controlled latency for real-time processing.

Unlike discrete encoding models \cite{zhang2023speechgpt,zhan-etal-2024-anygpt}, which require expanding the LLM vocabulary to support speech-text multimodality, we propose a continuous encoding-based speech LLM. In this framework, the output features of the streaming audio encoder are mapped to the LLM space through an MLP projection layer, enabling end-to-end speech understanding and generation. This design is inspired by LLaVA \cite{liu2024visual} but has been specifically optimized for streaming speech tasks.

\subsection{Data Format}

\begin{figure*}[ht]
    \centering
    \includegraphics[width=\linewidth]{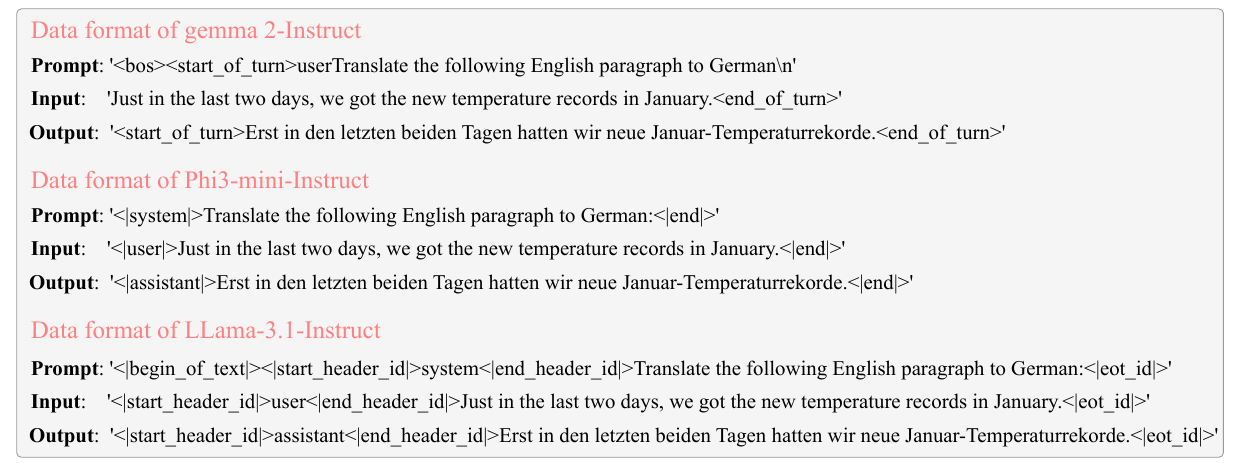}
    \caption{Data format of text translation task. An example of translation from English to German.}
    \label{Fig:Appendix-data-txt}
\end{figure*}

\begin{figure*}[ht]
    \centering
    \includegraphics[width=\linewidth]{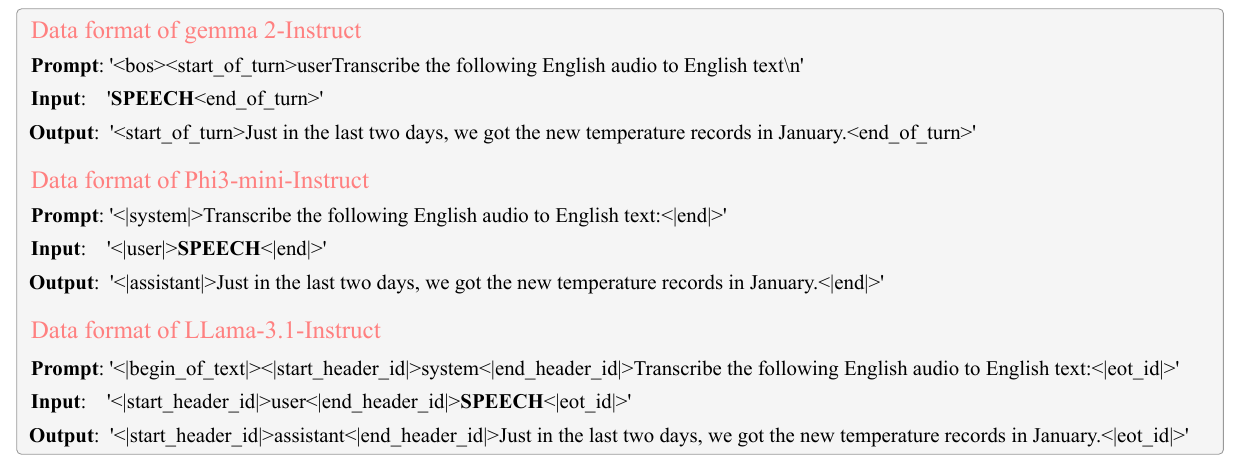}
    \caption{Data format of ASR task, where the '\textbf{SPEECH}' is the audio embedding for input.}
    \label{Fig:Appendix-data-speech}
\end{figure*}

In this paper, all the large language models we selected are instruction-tuned versions. To fully leverage their instruction-following capabilities, we strictly adhere to the instruction format used during their pretraining phase. Additionally, we design the data format, as shown in Figure~\ref{Fig:Appendix-data-txt} and Figure~\ref{Fig:Appendix-data-speech}, to align with the specific requirements of our tasks.

% \paragraph{Streaming ASR}

\subsection{Training Method}\label{Appendix-B3}
\paragraph{Training Method of Different Streaming Paradigms}
\begin{figure*}[ht]
    \centering
    \includegraphics[width=\linewidth]{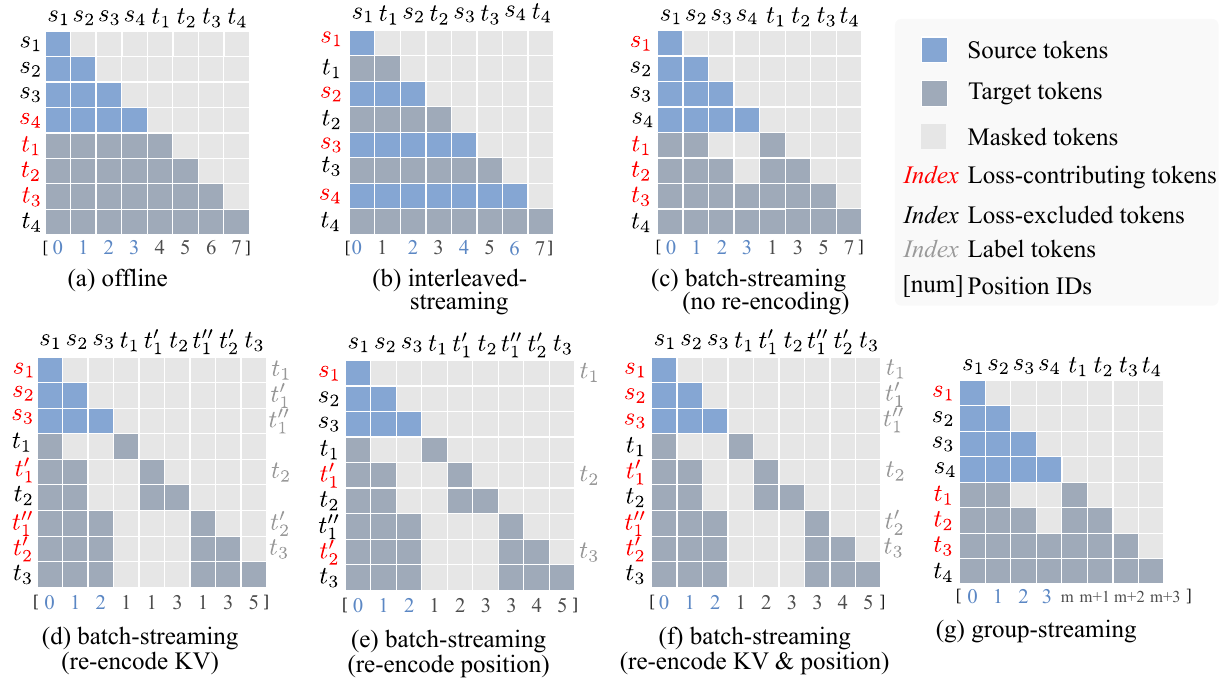}
    \caption{Attention mask matrix of different paradigms.}
    \label{Fig:Appendix-attn}
\end{figure*}

The main text analyzes the impact of different LLM paradigms on streaming tasks, covering training and evaluation methods for interleaved-streaming, batch-streaming, and group-streaming. This section provides a detailed explanation of the masking matrix design for different streaming paradigms and introduces the corresponding training methods. We explain these training paradigms in the context of the wait-k reading/writing policy \cite{ma2019stacl}.

Figure~\ref{Fig:Appendix-attn} illustrates the attention mask under different training methods, indicating the input order, position IDs, and whether a token is included in the loss calculation.
Figure~\ref{Fig:Appendix-attn} (a) represents the standard LLM causal mask matrix, which enables batch-processing training in offline scenarios using shifted loss computation.
Figure~\ref{Fig:Appendix-attn} (b) also employs a causal mask matrix, but the model's input consists of an interleaved sequence of source and target tokens, with position IDs assigned sequentially. Notably, each word may correspond to multiple tokens, where the first token is generated from the source, while the remaining tokens are generated from the target. During loss computation, only positions that contribute to target token generation are considered.
Figure~\ref{Fig:Appendix-attn} (c) depicts the batch-streaming mask matrix, which is structurally akin to \cite{raffel-etal-2024-simultaneous}. 
It maintains the batch-processing input format while adopting interleaved-streaming position encoding, preventing source tokens from attending to target tokens to eliminate input-attention mismatch and ensure streaming consistency.

Figures~\ref{Fig:Appendix-attn} (d), (e), and (f) represent three different re-encoding scenarios, all of which share the same mask matrix.
The core assumption of re-encoding is that as new content is continuously read at the source end, both the historical KV cache and position embedding must be updated accordingly to ensure accurate next-token prediction. Therefore, the training phase should reflect this dynamic updating mechanism. However, existing approaches employ either a causal-mask or prefix-to-prefix training methods \cite{raffel-etal-2024-simultaneous}, leading to a mismatch between training and inference. Specifically, causal-masked training is inherently offline and fails to capture the continuous update of content. While prefix-to-prefix training partially simulates this process, the target token is always the most recent one at each step. As more content is read, its behavior increasingly resembles an offline setting. In streaming scenarios, however, previously generated content cannot be modified, making this approach inadequate for capturing the true nature of re-encoding.
To address this discrepancy, the mask matrix design in Figures~\ref{Fig:Appendix-attn} (d), (e), and (f) ensures consistency between the training and inference processes, effectively aligning the training paradigm with real-world inference dynamics.

\paragraph{Training Method of Streaming ASR Model}
Due to the lack of a large-scale pre-trained streaming audio encoder, our modified streaming version of Wav2vec2 requires a step-by-step training approach. We adopt a four-stage training strategy to effectively train our proposed speech large language model, ensuring a smooth adaptation to streaming scenarios:
\begin{enumerate}
    \item \textbf{Stage 1: Pre-training for Feature Alignment.} In the first stage, we focus on establishing a robust feature alignment between the streaming audio encoder and the LLM. We begin by freezing both Wav2vec2 and the LLM and train the MLP projector using a batch-processing task. The goal is to learn a stable feature transformation that maps the continuous speech representations from Wav2vec2 into a space that aligns with the LLM's token embedding space. This step is crucial for minimizing the modality gap between speech and text representations, ensuring that the LLM can effectively process speech-derived embeddings in later stages.
    \item \textbf{Stage 2: Streaming Adaptation of Wav2vec2.} We replace Wav2vec2’s ConvPE with the causal version used in Wav2vec2-Streaming, enabling directional constraints suitable for streaming processing. In this stage, we jointly train Wav2vec2-Streaming and the projector, allowing the model to adapt to incremental encoding while maintaining alignment with the LLM’s input space.
    \item \textbf{Stage 3: Streaming Adaptation of Wav2vec2.} We replace Wav2vec2’s transformer encoder with the transformer decoder from Wav2vec2-Streaming. This modification ensures that the model adheres to global unidirectional constraints. We then continue joint training of Wav2vec2-Streaming and the projector, improving the encoder’s ability to generate high-quality speech embeddings in real-time.
    \item \textbf{Stage 4: Fine-tuning the LLM for Streaming ASR.} In the final stage, we freeze both Wav2vec2-Streaming and the projector, and fine-tune the LLM on a streaming ASR task. This step refines the LLM’s ability to generate accurate text outputs from streaming speech representations, optimizing its instruction-following capabilities while maintaining low-latency processing.
\end{enumerate}

\section{Experiments Details}
\subsection{Hyperparameters}
When verifying grouped position encoding, the model parameters are configured as shown in Table ~\ref{Table:App-para}. 
Notably, re-encoding introduces quadratic complexity, increasing the computational cost and resource requirements for both model training and inference. 
For the mismatch validation experiment, we reduce both the batch size and learning rate by half.

\begin{table}[h]
\centering
\begin{tabular}{lccc} 
\hline
Hyperparameter & \begin{tabular}[c]{@{}c@{}}Text to Text\\(Gemma2, Phi3, LLama3.1)\end{tabular} & \begin{tabular}[c]{@{}c@{}}ASR, Stage 1\\(Phi3)\end{tabular}& \begin{tabular}[c]{@{}c@{}}ASR, Stage 2 to 4\\(Phi3)\end{tabular}  \\ 
\hline
Precision      & bfloat16     & bfloat16  & bfloat16    \\
Learning Rate  & 2e-4         & 2e-4      & 2e-4        \\
LR Scheduler   & Linear       & Linear    & Linear      \\
Optimizer      & AdamW        & AdamW     & AdamW       \\
Warmup steps   & 500          & 1000      & 5000        \\
Lora rank      & 32           & 64        & 64          \\
Epochs         & 2            & 4         & 4           \\
Batch size     & 64           & 32        & 64          \\
Wait-k         & 1,3,5,7,9,11 & 1,3,5,7,9 & 1,3,5,7,9    \\
\hline
\end{tabular}
\caption{Fine-tuning hyperparameters of LLMs in this paper.}
\label{Table:App-para}
\end{table}

\subsection{Decoding Strategy}
The decoding process for streaming LLM is detailed in Algorithm~\ref{alg:decoding}.

\noindent\begin{minipage}{\textwidth}
\hfill
% \begin{minipage}{0.5\textwidth}
    \begin{algorithm}[H]
    % \label{alg:decoding}
    \caption{Streaming decoding with wait-k policy}
    \textbf{Input:} \parbox[t]{0.8\linewidth}{Source length list $S$, target length list $T$, wait-k policy $k$.} 
    \begin{algorithmic}[1]
    \State \parbox[t]{\linewidth}{Initialize source KV cache $S_{cache}$, target KV cache $T_{cache}$, and past token KV cache $P_{cache}$ as None.}
    \State \parbox[t]{\linewidth}{Initialize \textit{action=read}, ~\textit{is\_finished=false}, and $index=0$.}
    \State \parbox[t]{\linewidth}{Initialize~\textit{next\_token} as the target prompt tokens, and initialize generated tokens for this round \textit{token\_list} as an empty list.}
    \While{\textit{is\_finished} is \textit{false}}:
        \State \textbf{if} \textit{action} is \textit{read}:
            \State \qquad Separate $P_{cache}$ to $S_{cache}$ and $T_{cache}$.
            \State \qquad Read prompt and $k+index$ words, and save hidden state to source KV cache $S_{cache}$.
            \State \qquad Merge $S_{cache}$ and $T_{cache}$ to $P_{cache}$.
            \State \qquad Set \textit{action=write}.
            \State \qquad Set $index = index + 1$.
        \State \textbf{elif} \textit{action} is \textit{write}:
            \State \qquad Separate $P_{cache}$ to $S_{cache}$ and $T_{cache}$.
            \State \qquad Calculate and save hidden state to target KV cache $T_{cache}$.
            \State \qquad Merge $S_{cache}$ and $T_{cache}$ to $P_{cache}$.
            \State \qquad Project \textit{next\_token} as Q, and calculate attention output with  KV cache $P_{cache}$.
            \State \qquad Predict and update the \textit{next\_token} based on greedy decoding.
            \State \qquad \textbf{if} \textit{next\_token} is the end symbol:
            \State \qquad \qquad Set \textit{is\_finished} as \textit{true}.
            \State \qquad Add \textit{next\_token} to \textit{token\_list}.
            \State \qquad \textbf{if} \textit{token\_list} forms a complete word:
            \State \qquad \qquad \textbf{Print} the word.
            \State \qquad \qquad Set \textit{action} as \textit{read}, reset \textit{token\_list} as an empty list.
    \EndWhile
    \State \textbf{Return:} The predict words.

    \end{algorithmic}
    \label{alg:decoding}
    \end{algorithm}
\end{minipage}

\section{More Details about Group Position}\label{Appendix-E}
\subsection{Relative Distance of Group Position}
Let the source tokens be $X=[x_0,x_1,\ldots,x_{M-1}]$ and target tokens be $Y=[y_0,y_1,\ldots,y_{N-1}]$, where the position IDs are $pos_x=[0,1,\ldots,M-1]$ and $pos_y=[\phi,\phi+1,\ldots,\phi+N-1]$, respectively.
In batch-processing mode, the starting position ID on the target side is given by $\phi=M$.
In batch-streaming mode, the starting position ID on the target side is given by $\phi=0$.

Define the rotary matrix as $R(m) =diag(R_1(m),R_2(m),\ldots,R_{d/2}(m))$, where $m$ is the position id, $d$ is the model dimension, and 
\begin{equation}
    R_i(m)=\begin{bmatrix}\cos(m\theta_i)&-\sin(m\theta_i)\\\sin(m\theta_i)&\cos(m\theta_i)\end{bmatrix}, \quad \theta_i=10000^{-2i/d}.
\end{equation}

For the original rotary position embedding (RoPE) \cite{su2024roformer}, positional information is incorporated into each token’s Query ($q$) and Key ($k$) through a rotation matrix. This process can be expressed as $q^r_n = R(n)q_n$ and $k^r_m = R(m)k_m$, where $n$ and $m$ denote the respective position IDs.
Then, the attention mechanism in RoPE incorporates the rotationally transformed queries and keys, leading to the attention score computation as follows:
\begin{equation}
    Attention(n,m)={q^r_n}^Tk^r_m=q_n^TR^T(n)R(m)k_m=q_n^TR(m-n)k_m.
\end{equation}

For any two positions $n$ and $m$ within the sequence, their position encoding depends solely on $R(m-n)$, meaning it is determined by their relative distance $m-n$. 
When $k_m$ and $q_n$ both belong to either source tokens or target tokens, the relative distance is given by $\Delta = m - n$. In this case, the positional encoding results in batch-processing and batch-streaming remain identical.
When $k_m$ and $q_n$ belong to source tokens and target tokens, respectively, the relative distance is given by $\Delta = \phi + j - m$, where $j$ denotes the position of $q_n$ as the $j$-th token on the target side. In this case, the positional encoding results in batch-processing and batch-streaming depend on $\phi$.

\begin{figure*}[t]
    \centering
    \includegraphics[width=0.95\linewidth]{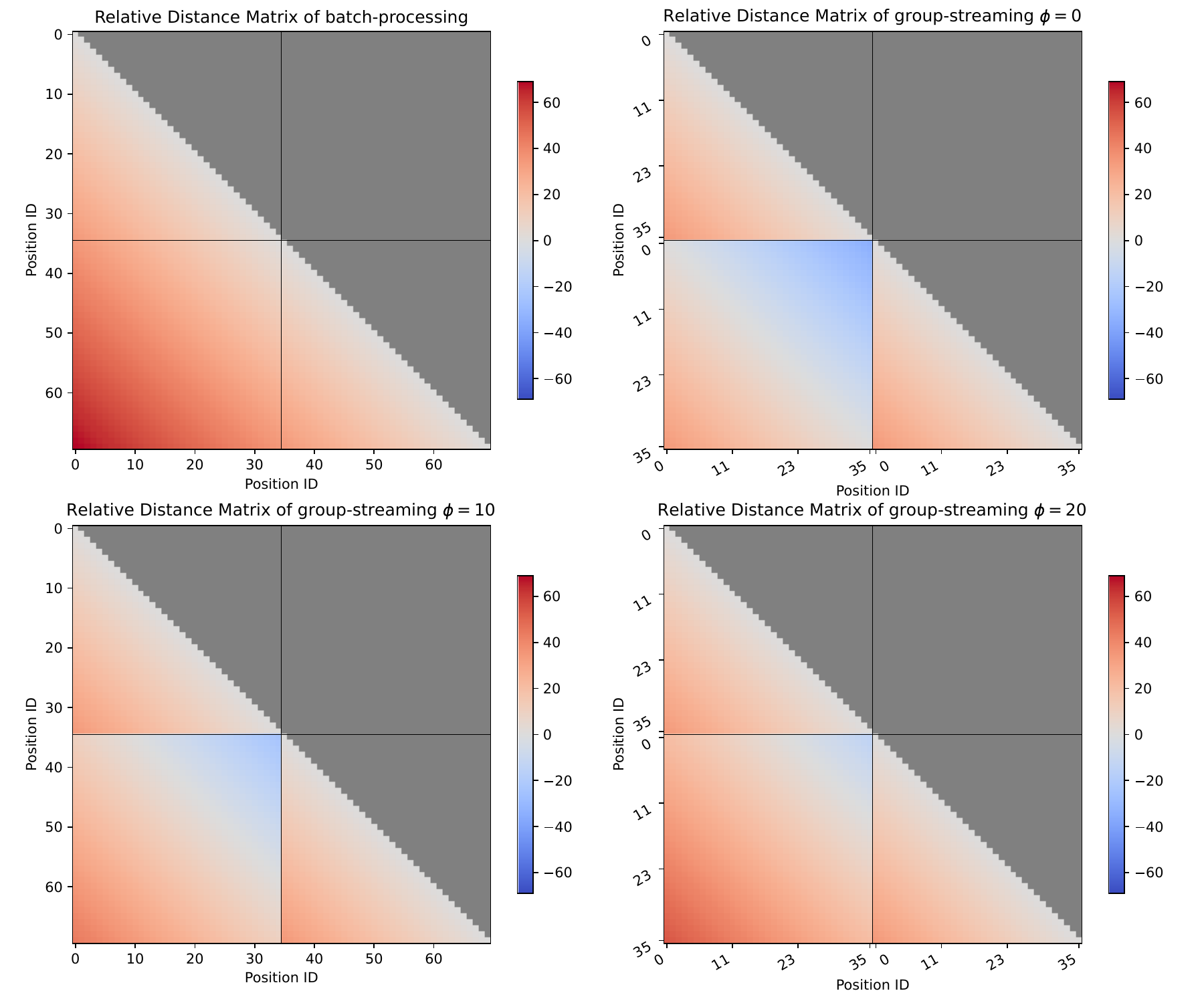}
    \caption{Relative distance matrix of batch-processing mode and group-streaming mode.}
    \label{Fig:Pos}
\end{figure*}

For batch processing, $\phi = M - 1$ indicates that the target tokens are farther from the source starting token and closer to the source ending token.\textbf{ In contrast, for batch-streaming, $\phi = 0$ means the target starting token is closer to the source starting token and farther from the source ending token.} This aligns with the sequential information arrival order in streaming scenarios, making it more suitable for capturing relative positional changes in streaming settings.

Figure~\ref{Fig:Pos} illustrates the variation in relative distances under batch-processing and batch-streaming settings. In batch-processing mode, which is typically used in offline scenarios, position IDs are assigned sequentially. Tokens near the diagonal exhibit local positional relationships with smaller relative distances, whereas tokens farther from the diagonal have increasingly larger relative distances, reflecting their positional separation.  
In batch-streaming mode, the relative positional relationships among tokens within the source and target sequences remain unchanged. However, the relative distance between target and source tokens is influenced by the parameter $\phi$, shifting accordingly as $\phi$ increases. Taking $\phi = 0$ as an example, in a streaming scenario, the target token with position ID 0 first interacts with the source token at position ID 0, resulting in a relative distance of 0. This alignment effectively models the sequential nature of data accumulation in streaming settings, ensuring that the position encoding adapts dynamically to the progressive arrival of information.

\subsection{Why Group Position Avoids Confusion}
Research by \cite{shen2018disan,haviv2022transformer,tsai2019transformer} have shown that decoder-only models can learn implicit positional information. In decoder-only architectures, source tokens and target tokens attend to different contexts. As a result, even if their position IDs overlap, the model can still distinguish between source and target based on the content they attend to. As shown in Equation \ref{eq_confuse}:
\begin{align}
Attention(n,cache)&=\sum _{i=0}{q^r_n}^T{k_i}^r=\sum _{i=0}q_n^TR^T(n)R(i)k_i=\sum _{i=1}q_n^TR(i-n)k_i \notag\\
&= 
\begin{cases}
\sum_{i=0}^{n}q_{n_s}^TR(i-n)k_{i_s}, & q_{n_s}~is~source, \\
\\
\sum_{i=0}^{M-1}q_{n_t}^TR(i-n)k_{i_s} + \sum _{i=0}^{n}q_{n_t}^TR(i-n)k_{i_t}, & q_{n_t}~is~target.
\end{cases}
\label{eq_confuse}
\end{align}
% where 
In both cases, the query token has an ID of $n$, but since it attends to different content, the model can still distinguish between the source and the target.

% \begin{align}
% Attention(n,cache)&=\sum _{i=0}{q^r_n}^T{k_i}^r=\sum _{i=0}q_n^TR^T(n)R(i)k_i=\sum _{i=1}q_n^TR(i-n)k_i\\
% &= \sum _{i=0}^{M-1}q_n^TR(i-n)k_{i_s} + \sum _{i=0}^{n}q_n^TR(i-n)k_{i_t}
% \end{align}

% When $m<n<M$ or $M-1<m<n$, the positional encodings in both modes remain identical.

% \Delta

\subsection{Potential Edge of Group Position}\label{Appendix-D3}
The model can learn the position offset $\phi$ through simple fine-tuning, so typical values of $\phi$ do not significantly impact performance. However, when $\phi$ becomes extremely large, it may lead to discrepancies with the model’s pretraining distribution due to the limited context length used during pretraining. Therefore, a reasonable range for $\phi$ should ensure that the maximum relative distance, specifically, between the last target token and the first source token, does not exceed the model’s pretraining context length. For example, Gemma2-2B-Instruct was pretrained with a context length of 8k, which suggests that the maximum suitable value of $\phi$ is around 6k, as shown in Table~\ref{tab:gem ma-memory-waitk}.

\begin{table}[h]
\centering
\footnotesize
\setlength{\tabcolsep}{6pt}
\renewcommand{\arraystretch}{1.2} 
\begin{tabular}{@{}lccccccccccc@{}}
\toprule
\textbf{Model} & \textbf{Wait-k} & \textbf{m=0} & \textbf{m=512} & \textbf{m=4k} & \textbf{m=5k} & \textbf{m=6k} & \textbf{m=7k} & \textbf{m=8k} & \textbf{m=10k} & \textbf{m=50k} \\
\midrule
\multirow{2}{*}{Gemma2-2B-Instruct (8k)} 
& 5 & \textbf{40.76} & 40.68 & 40.70 & 40.51 & \textbf{40.21} & 39.83 & 39.73 & 39.52 & 39.37 \\
& 9 & \textbf{40.91} & 40.89 & 40.85 & 40.81 & \textbf{40.73} & 40.11 & 39.97 & 39.78 & 39.56 \\
\bottomrule
\end{tabular}
\caption{BLEU performance of Gemma2-2B-Instruct (8k) under different memory sizes $m$ and wait-$k$ settings.}
\label{tab:gem ma-memory-waitk}
\end{table}

% \section{Additional Results}
% \vspace{-1\baselineskip}
\section{Full Results of Text Translation Task}\label{Appendix-D}
This section provides additional results to validate the impact of different initial position IDs on the target side in streaming translation tasks. The results cover three different large language models and two different translation tasks.
The full results of the text translation task are shown in Table~\ref{tab:bleu-al}, which includes the accuracy metric BLEU and the latency metric LAAL.
\begin{table*}[ht]
    \centering
    \footnotesize
    \setlength{\tabcolsep}{7pt}
    \renewcommand{\arraystretch}{1.15} 
    \begin{tabular}{c|c|llllll} 
    \toprule
    \multirow{2}{*}{\textbf{Dataset}} & \multirow{2}{*}{\textbf{Wait-k}}          & \multicolumn{6}{c}{\textbf{\textbf{Gemma2-2b-Instruct~\textbf{\textbf{~}}}}(Target start id $\phi$)}                                                                                                              \\ 
    \cmidrule(l){3-8}
                                      &                                           & \multicolumn{1}{c}{0}             & \multicolumn{1}{c}{0.5}           & \multicolumn{1}{c}{128}           & \multicolumn{1}{c}{256}           & \multicolumn{1}{c}{512}           & \multicolumn{1}{c}{$\Delta$}  \\ 
    \cmidrule{1-8}
    \multirow{4}{*}{En-Fr}            & 5                                         & 40.76~(5.21)                      & 40.76~(5.21)                      & 40.70 (5.21)                      & 40.57 (5.20)                      & 40.68 (5.21)                      & \cellcolor{gray!20} 0.19 (0.01)                                    \\
                                      & 7                                         & 40.92 (7.18)                      & 40.92 (7.18)                      & 40.85 (7.17)                      & 40.91~(7.18)                      & 40.92~(7.18)                      & \cellcolor{gray!20} 0.07 (0.01)                                    \\
                                      & 9                                         & 40.91 (9.14)                      & 40.91 (9.14)                      & 40.90 (9.13)                      & 40.88 (9.13)                      & 41.01 (9.13)                      & \cellcolor{gray!20} 0.09 (0.01)                                    \\
                                      & 11                                        & 41.10 (11.09)                     & 41.10 (11.09)                     & 41.14 (11.09)                     & 40.96 (11.09)                     & 41.05 (11.09)                     & \cellcolor{gray!20} 0.18 (0.00)                                    \\ 
    \cline{2-8}
    \multirow{4}{*}{En-De}            & 5                                         & 30.84~(4.62)                      & 30.84~(4.62)                      & 30.90~(4.63)                      & 30.80~(4.62)                      & 30.95~(4.56)                      & \cellcolor{gray!20} 0.15 (\underline{0.07})                                    \\
                                      & 7                                         & 31.47 (6.63)                      & 31.47~(6.63)                      & 31.44~(6.63)                      & 31.57~(6.63)                      & 31.67~(6.59)                      & \cellcolor{gray!20} \underline{0.23} (0.04)                                    \\
                                      & 9                                         & 31.73~(8.66)                      & 31.73~(8.66)                      & 31.87~(8.65)                      & 31.91~(8.65)                      & 31.88~(8.65)                      & \cellcolor{gray!20} 0.18 (0.01)                                    \\
                                      & 11                                        & 31.95~(10.70)                     & 31.95~(10.70)                     & 31.98~(10.69)                     & 31.95~(10.69)                     & 31.89~(10.69)                     & \cellcolor{gray!20} 0.09 (0.01)                                    \\ 
    \hline
    \multirow{2}{*}{\textbf{Dataset}} & \multirow{2}{*}{\textbf{\textbf{Wait-k}}} & \multicolumn{6}{c}{\textbf{\textbf{Phi3-mini-Instruct~}}(Target start id $\phi$)}                                                                                                                                 \\ 
    \cmidrule(l){3-8}
                                      &                                           & \multicolumn{1}{c}{0}             & \multicolumn{1}{c}{0.5}           & \multicolumn{1}{c}{128}           & \multicolumn{1}{c}{256}           & \multicolumn{1}{c}{512}           & \multicolumn{1}{c}{$\Delta$}  \\ 
    \cmidrule{1-8}
    \multirow{4}{*}{En-Fr}            & 5                                         & 39.89 (5.45)                      & 39.89 (5.45)                      & 39.91 (5.44)                      & 40.06 (5.41)                      & 39.87 (5.44)                      &\cellcolor{gray!20} 0.19 (0.03)                                    \\
                                      & 7                                         & 40.57 (7.38)                      & 40.57~(7.38)                      & 40.53~(7.37)                      & 40.72~(7.39)                      & 40.71~(7.39)                      & \cellcolor{gray!20} \cellcolor{gray!20} 0.19 (0.02)                                    \\
                                      & 9                                         & 41.31 (9.28)                      & 41.31 (9.28)                      & 41.04 (9.29)                      & 41.35 (9.27)                      & 41.44~(9.27)                      & \cellcolor{gray!20} 0.20 (0.02)                                    \\
                                      & 11                                        & 41.92 (11.17)                     & 41.92 (11.17)                     & 42.03 (11.17)                     & 41.94 (11.17)                     & 41.93 (11.17)                     & \cellcolor{gray!20} 0.11 (0.00)                                    \\ 
    \cline{2-8}
    \multirow{4}{*}{En-De}            & 5                                         & 30.92 (4.65)                      & 30.92~(4.65)                      & 30.76~(4.64)                      & 30.81~(4.65)                      & 30.86~(4.65)                      & \cellcolor{gray!20} 0.16 (0.01)                                    \\
                                      & 7                                         & 31.94 (6.65)                      & 31.94~(6.65)                      & 31.78~(6.64)                      & 31.84~(6.64)                      & 31.78 (6.64)                      & \cellcolor{gray!20} 0.16 (0.01)                                    \\
                                      & 9                                         & 32.18~(8.69)                      & 32.18~(8.69)                      & 32.10~(8.68)                      & 32.21~(8.69)                      & 32.09~(8.68)                      & \cellcolor{gray!20} 0.12 (0.01)                                    \\
                                      & 11                                        & 32.26 (10.71)                     & 32.26 (10.71)                     & 32.23~(10.73)                     & 32.23~(10.73)                     & 32.28~(10.73)                     & \cellcolor{gray!20} 0.10 (0.02)                                    \\ 
    \hline
    \multirow{2}{*}{\textbf{Dataset}} & \multirow{2}{*}{\textbf{\textbf{Wait-k}}} & \multicolumn{6}{c}{\textbf{\textbf{LLaMA3.1-8b-Instruct~}\textbf{~}}(Target start id $\phi$)}                                                                                                                     \\ 
    \cmidrule(l){3-8}
                                      &                                           & \multicolumn{1}{c}{0}             & \multicolumn{1}{c}{0.5}           & \multicolumn{1}{c}{128}           & \multicolumn{1}{c}{256}           & \multicolumn{1}{c}{512}           & \multicolumn{1}{c}{$\Delta$}  \\ 
    \cmidrule{1-8}
    \multirow{4}{*}{En-Fr}            & 5                                         & 40.11~(5.23)          & 40.11~(5.23)            & 40.10 (5.22)            & 39.93~(5.23)            & 39.92~(5.23)            & \cellcolor{gray!20} 0.19 (0.01)                                    \\
                                      & 7                                         & 40.30 (7.19)          & 40.30~(7.19)            & 40.32~(7.19)            & 40.35~(7.20)            & 40.31~(7.19)            & \cellcolor{gray!20} \textbf{0.03} (0.01)                                    \\
                                      & 9                                         & 40.15 (9.17)          & 40.15~(9.17)            & 40.32~(9.16)            & 40.34~(9.17)            & 40.35~(9.17)            & \cellcolor{gray!20} 0.20 (0.01)                                    \\
                                      & 11                                        & 40.53~(11.11)         & 40.53~(11.11)           & 40.47~(11.11)           & 40.58~(11.11)           & 40.63~(11.10)           & \cellcolor{gray!20} 0.16 (0.01)                                    \\ 
    \cline{2-8}
    \multirow{4}{*}{En-De}            & 5                                         & 30.33 (4.58)                      & 30.33~(4.58)                      & 30.21~(4.57)                      & 30.37~(4.58)                      & 30.34~(4.58)                      & \cellcolor{gray!20} 0.16 (0.01)                                    \\
                                      & 7                                         & 31.23 (6.54)                      & 31.23 (6.54)                      & 31.18~(6.54)                      & 31.16~(6.54)                      & 31.25~(6.53)                      & \cellcolor{gray!20} 0.09 (0.01)                                    \\
                                      & 9                                         & 31.80 (8.63)                      & 31.80~(8.63)                      & 31.83~(8.63)                      & 31.76~(8.62)                      & 31.89 (8.62)                      & \cellcolor{gray!20} 0.13 (0.01)                                    \\
                                      & 11                                        & 32.04 (10.56)                     & 32.04 (10.56)                     & 31.98~~(10.55)                    & 32.07~(10.56)                     & 32.08~(10.56)                     & \cellcolor{gray!20} 0.10 (\textbf{0.00})                                    \\
    \bottomrule
    \end{tabular}
    \caption{Performance comparison of different models with various wait-k policies and target start IDs. $\Delta$ represents the range of variation in BLEU scores and LAAL scores when the start id of target token takes different values. We use bold to indicate the smallest variation. Underline represents the largest variation.}
    \label{tab:bleu-al}
\end{table*}

% \subsection{Zero-shot Performance of LLMs on Streaming Tasks}

\section{Model Efficiency}
% To better understand the efficiency advantages of the proposed grouped-streaming framework, this section compares the computational cost and throughput between re-encoding and our grouped-streaming approach. In practical streaming scenarios, the cost of repeatedly encoding overlapping source contexts can become a major bottleneck. By removing this redundancy, our grouped paradigm is expected to significantly reduce latency and improve throughput.
This section compares the computational cost and throughput between re-encoding and our grouped-streaming approach.
We conduct a case study on the En–Fr streaming translation task using a filtered subset of the dataset that contains 7.3k sentence-level examples with controlled lengths, amounting to approximately 32k tokens.All experiments are conducted using the Phi-3 Mini Instruct model. Table~\ref{tab:waitk-efficiency-final} summarizes the inference time and throughput under different Wait-k settings, highlighting the drastic efficiency gains brought by removing re-encoding.

\begin{table}[b]
\centering
\footnotesize
\setlength{\tabcolsep}{10pt}
\renewcommand{\arraystretch}{1.0}
\begin{tabular}{c l cl cc}
\toprule
{Wait-k} & {Inference mode} & {Time consumption} && {Throughput} &\\
\midrule
\multirow{2}{*}{5} 
& with re-encoding & 59.54 h && 1.79 tokens/s& \\
& \textbf{without re-encoding} & \textbf{4.38 h} &$\downarrow$ \textbf{92.6\%} & \textbf{20.24 tokens/s} &$\times$ \textbf{11.3} \\
\midrule
\multirow{2}{*}{9} 
& with re-encoding & 28.87 h && 3.70 tokens/s &\\
& \textbf{without re-encoding} & \textbf{4.04 h} &$\downarrow$ \textbf{86.1\%} & \textbf{21.93 tokens/s}& $\times$ \textbf{5.9} \\
\bottomrule
\end{tabular}
\caption{Comparison of inference efficiency under different Wait-k values and re-encoding modes.}
\label{tab:waitk-efficiency-final}
\end{table}

The results in the figure show that the proposed grouped-streaming paradigm eliminates the need for re-encoding, resulting in significant throughput improvements: over 5 $\times$ speedup under the wait-9 setting and more than 11 $\times$ speedup under wait-5 setting, compared to the re-encoding baseline.

\section{Visualization}

\vspace{-0.5\baselineskip}
\subsection{Attention Distribution}\label{Appendix-F-1}
\begin{figure*}[ht]
    \centering
    \includegraphics[width=\linewidth]{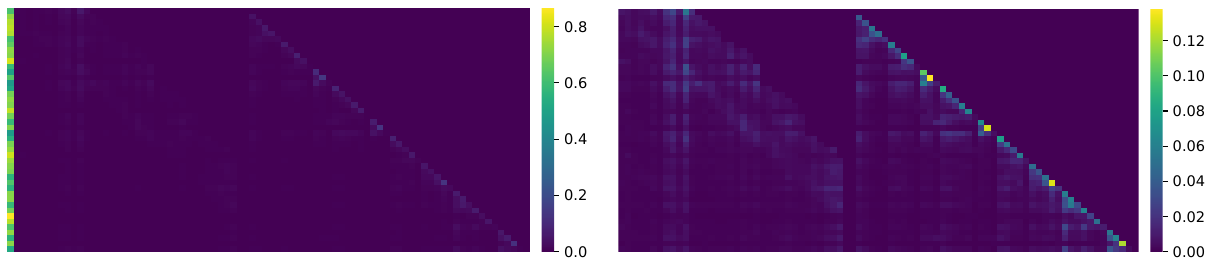}
    \caption{The absolute values of the attention matrix, with the left figure incorporating the attention sink, while the right figure depicts the matrix after the removal of the attention sink.}
    \label{Fig:Appendix-Attn-pre}
\end{figure*}

Figure~\ref{Fig:Appendix-Attn-pre} illustrates the absolute values of the attention matrix, representing the attention magnitude of target tokens to both the input and output. In the left figure, the most attended column corresponds to the attention sink \cite{xiaoefficient}, whereas in the right figure, the attention sink has been removed.
The absolute attention map highlights each token's attention to historical tokens but makes it difficult to assess how different tokens distribute their attention toward a specific token.
To better emphasize the distribution of target tokens' attention toward a given token, we normalize the attention map along columns and apply a gamma transformation to enhance and amplify the relationships. Mathematically, given an attention matrix $A$, where $A_{i,j}$  represents the attention weight from token $j$ to token $i$, we normalize each column as follows:
\vspace{-0.5\baselineskip}
\begin{equation}
    A_{i,j}^{\prime}=\Bigg(\frac{A_{i,j}-\min_i\{A_{i,j}\}}{\max_i\{A_{i,j}\}-\min_i\{A_{i,j}\}}\Bigg).
\end{equation}
% where we set $\gamma$ as 0.5.

\vspace{-0.5\baselineskip}

\subsection{Example of Streaming Decoding}
\begin{figure*}[!ht]
    \centering
    \includegraphics[width=\linewidth]{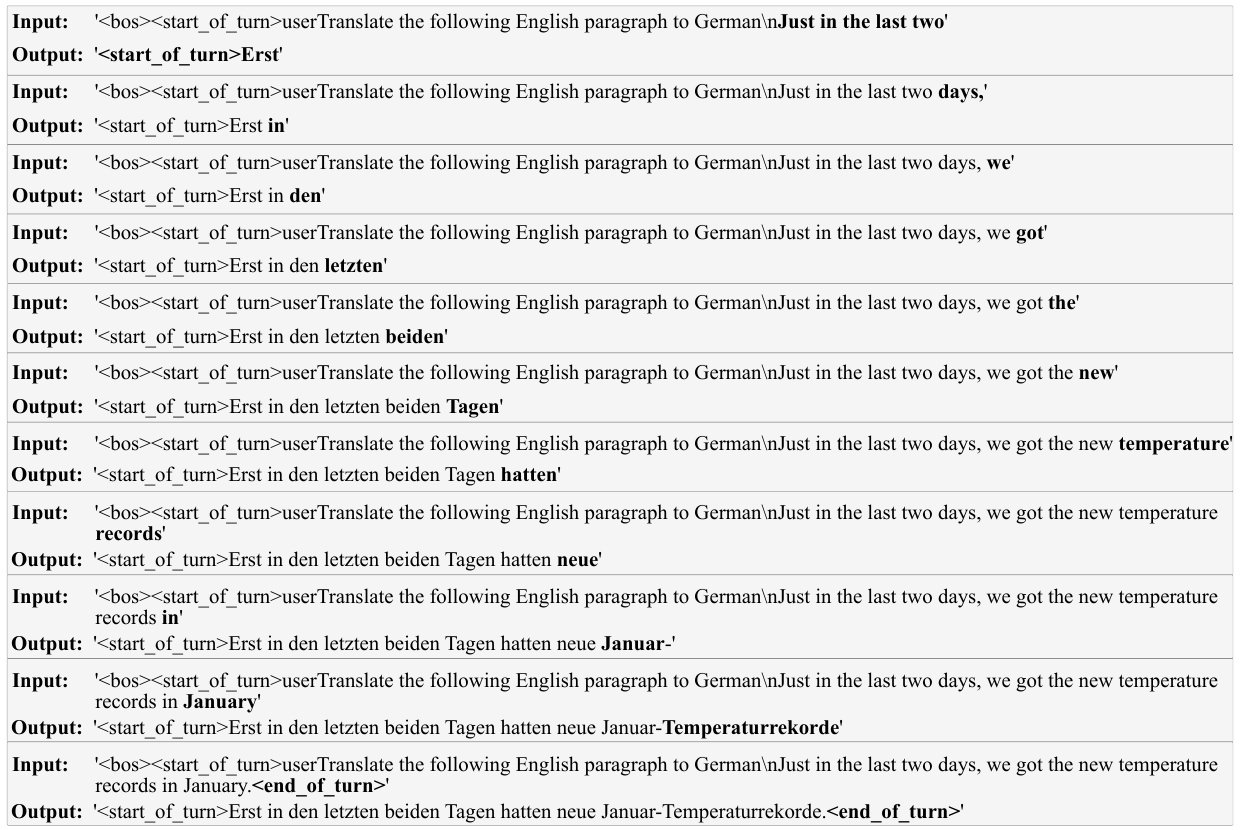}
    % \caption{An English to German translation example of streaming decoding with wait-5 reading/writing policy. The bold words indicate the most recently read or generated content.}
    % \caption{English to German streaming translation. The bold indicate the most recently read or generated content.}
    \caption{An example on wait-5 reading/writing policy. The bold indicate the most recently content.}
    \label{Fig:Appendix-decoding-example}
\end{figure*}

Figure~\ref{Fig:Appendix-decoding-example} is an example of streaming reading and decoding process.
\vspace{-0.5\baselineskip}

\end{document}